\documentclass[letterpaper]{article} 
\usepackage{url}  
\usepackage{graphicx}  
\usepackage[margin=1in]{geometry}

\usepackage{booktabs}       
\usepackage{amsfonts}       
\usepackage{nicefrac}       
\usepackage{microtype}      

\newcommand{\citet}[1]
{\citeauthor{#1} ~\shortcite{#1}}
\newcommand{\citep}{\cite}
\newcommand{\citealp}[1]
{\citeauthor{#1} ~\citeyear{#1}}

\usepackage{algorithm}
\usepackage{algorithmic}

\usepackage{subfigure} 
\graphicspath{{Figures/}}

\usepackage{amsmath,amsthm}
\usepackage{amssymb}
\usepackage{mathrsfs}
\DeclareMathOperator*{\argmin}{argmin}

\usepackage{hyperref}
\hypersetup{backref,colorlinks=true,citecolor=blue,linkcolor=blue,urlcolor=blue}
\usepackage[numbers,square,sort]{natbib}

\newtheorem{theorem}{Theorem}

\newtheorem{lemma}[theorem]{Lemma}

\usepackage{gensymb}
\usepackage{xcolor}

\newcommand{\one}{\mathbf{1}}

\newcommand{\given}{\;\vert\;}
\newcommand{\norm}[1]{\left\lVert #1 \right\rVert}
\newcommand{\y}{\textrm{year}}
\DeclareMathOperator*{\prox}{\bf prox}
\DeclareMathOperator*{\sign}{sign}

\usepackage{mathtools}
\mathtoolsset{showonlyrefs}

\begin{document} 

\title{Algorithms for Estimating Trends in Global Temperature Volatility}

\author{Arash Khodadadi and Daniel J. McDonald\\
 Department of Statistics\\
 Indiana University\\
 Bloomington, IN 47408 \\
 \{arakhoda,dajmcdon\}@indiana.edu}

\maketitle

\begin{abstract}
Trends in terrestrial temperature variability are perhaps more relevant for species
viability than trends in mean
temperature. In this paper, we develop methodology for estimating such
trends using multi-resolution climate data from polar orbiting weather
satellites. We derive two novel algorithms for computation that are
tailored for dense, gridded observations over both 
space and time. We evaluate our methods with a simulation that mimics these data's 
features and on a large, publicly available, global
temperature dataset with the eventual goal of tracking trends in
cloud reflectance temperature variability.
\end{abstract}

\section{Introduction}


The amount of sunlight reflected from clouds is among the largest
sources of uncertainty in climate
prediction~\citep{BoucherRandall2013}. But climate models fail to
reproduce global cloud statistics, and understanding the reasons for
this failure is a grand challenge of the World Climate Research
Programme~\citep{BonyStevens2015}.
While numerous
studies have examined the overall impacts of clouds on climate
variability~\citep{MyersMechoso2018,GrisePolvani2013,BenderRamanathan2012},
such investigations have been hampered by the lack of a suitable
dataset. Ideal data would have global coverage at high spatial
resolution, a long enough record to recover temporal trends, and be
multispectral~\citep{WielickiYoung2013}. 
To address this gap, current
work~\cite{StatenKahn2016,SchreierKahn2010,KahnFishbein2007} seeks to
create a 
spectrally-detailed dataset by 
combining radiance data from Advanced Very High Resolution Radiometer
imagers with readings from High-resolution 
Infrared Radiation Sounders, instruments onboard legacy weather
satellites. 
In anticipation of this new dataset, our work develops novel methodology
for examining the trends in variability of climate data across space
and time.

\subsection{Variability Rather Than Average}
\label{sec:vari-rath-than}

Trends in terrestrial temperature variability are perhaps more relevant for species
viability than trends in mean
temperature~\cite{huntingford_no_2013},
because an increase in
temperature variability will increase the probability of extreme hot
or cold
outliers~\cite{VasseurDeLong2014}. Recent climate literature
suggests that it is more difficult for society to adapt to these
extremes than to the gradual increase in the mean temperature
\cite{hansen_perception_2012,huntingford_no_2013}. Furthermore, the willingness of popular media to
emphasize the prevalence extreme cold events coupled with a
fundamental misunderstanding of the relationship between climate (the global
distribution of weather over the long run) and weather (observed
short-term, localized behavior) leads to public misunderstanding of climate
change. In fact, a point of active debate is the extent to which the observed
increased frequency of extreme cold events in the 
northern hemisphere can be attributed to increases in temperature
variance  rather than to changes in
mean climate~\cite{Screen2014,FischerBeyerle2013,TrenberthZhang2014}.

Nevertheless, research examining trends in the volatility of
spatio-temporal climate data is scarce. \citet{hansen_perception_2012} studied the change in
the standard deviation (SD) of the surface temperature in the NASA
Goddard Institute for Space Studies gridded temperature dataset by
examining the empirical SD at each spatial location relative to that
location's SD over a base period and showed
that these estimates are increasing.
\citet{huntingford_no_2013} took a similar
approach in analyzing the ERA-40 data set. They argued that, while there
is an increase in the SDs from 1958-1970 to 1991-2001, it
is much smaller than found by
\citet{hansen_perception_2012}. \citet{huntingford_no_2013} also computed the
time-evolving global SD from the detrended time-series at each
position and argued that the global SD has been
stable. 

These and other related work (e.g., \citealp{rhines_frequent_2013}) have
several shortcomings which our work seeks to remedy. First, no 
statistical analysis has been performed to examine if the changes in
the SD are statistically significant. Second, the methodologies for
computing the SDs are highly sensitive to the choice of base period. Third,
and most importantly, temporal and spatial correlations between 
observations are completely ignored.  

Importantly, existing literature and our present work examines variance (rather
than the mean) for a number of reasons. 
First, instrument bias in the satellites increases over time so
examining the mean over time conflates that bias with any actual
change in mean (though the variance is unaffected). 
Second, extreme weather events (hurricanes, droughts, wildfires in
California, heatwaves in Europe) may be driven more strongly by increases
in variance than by increases in mean. 
Finally, even if the global mean temperature is constant, there may still
be climate change. In fact, atmospheric physics suggests that,
across space, average temperatures should not change (extreme cold
in one location is offset by heat in another). But if swings across
space are becoming more rapid, then, even with no change in mean global
temperature over time, increasing variance can lead to
increases in the prevalence of extreme events.

\subsection{Main Contributions}

The main contribution of this work is to develop a new methodology for
detecting the trend in the volatility of spatio-temporal data. In this
methodology, the variance at each position and time are estimated by
minimizing the penalized negative
loglikelihood. Following methods for mean estimation~\cite{Tibshirani2014}, we penalize the  
differences between the estimated variances which
are temporally and spatially ``close'', resulting
in a generalized LASSO problem. However, in our application, the
dimension of this optimization problem is massive, so the
standard solvers are inadequate. 

We develop two algorithms
which are computationally feasible on extremely large data. In the
first method, we adopt an optimization technique 
called alternating direction method of multipliers
(ADMM, \citealp{boyd_distributed_2011}), to divide the total problem into
several sub-problems of much lower dimension and show how the total
problem can be solved by iteratively solving these sub-problems. The
second method, called linearized ADMM
\cite{parikh_proximal_2014}, solves the main problem by iteratively
solving a linearized version. We will compare the benefits of
each method.

Our main contributions are as follows:
\begin{enumerate}
\item We propose a method for nonparametric variance estimation for a
  spatio-temporal process and discuss the relationship between our
  methods and those existing in the machine learning literature (\autoref{sec:ell_1-trend-filt}).
\item We derive two alternating direction method of multiplier
  algorithms to fit our estimator when applied to 
  very large data (\autoref{sec:prop-optim-meth}). We give situations
  under which each algorithm is most likely to be useful. Open-source
  Python code is available.\footnote{\tt github.com/dajmcdon/VolatilityTrend}
\item Because the construction of satellite-based datasets is ongoing
  and currently proprietary, we illustrate our methods on a large,
  publicly available, global temperature dataset.
  The goal is to demonstrate the feasibility of these methods for tracking
  world-wide trends in variance in standard atmospheric data and a
  simulation constructed to mimic these data's features
  (\autoref{sec:empirical-evaluation}). 
\end{enumerate}

While the motivation for our methodology is its application to large, gridded climate
data, we note that our algorithms are easily generalizable to
spatio-temporal data under convex loss, e.g. exponential family
likelihood. Furthermore the spatial structure can be broadly construed
to include general graph dependencies. Our current application uses
Gamma likelihood which lends itself well to modeling trends in
pollutant emissions or in astronomical phenomena like microwave
background radiation. Volatility estimation in oil and natural gas markets
or with financial data is another possibility. Our methods can also
be applied to resting-state fMRI data (though the penalty structure changes).

\section{Smooth Spatio-temporal Variance Estimation}
\label{sec:ell_1-trend-filt}

\citet{KimKoh2009} proposed $\ell_1$-trend filtering as a
method for estimating a smooth, time-varying trend. It is formulated
as the optimization problem 
$$
\min_{\beta} \frac{1}{2} \sum_{t=1}^{T} (y_t-\beta_t)^2+\lambda
\sum_{t=2}^{T-1} \left|\beta_{t-1}-2\beta_{t}+\beta_{t+1}\right| 
$$
or equivalently:
\begin{equation}
\min_{\beta} \frac{1}{2} \norm{ y-\beta }_2^2+\lambda \norm{ D_t \beta}_1
\label{eq:l1tf}
\end{equation}
 where $y=\{y_t\}_{t=1}^T$ is an observed time-series, $\beta\in \mathbb{R}^T$ is the smooth trend,
 $D_t$ is a $(T-2)\times T$ matrix, and $\lambda$ is a tuning parameter
 which balances fidelity to the data (small errors in the first term)
 with a desire for smoothness.  
\citet{KimKoh2009} proposed a specialized primal-dual
interior point (PDIP) algorithm for solving \eqref{eq:l1tf}. From a
statistical perspective, \eqref{eq:l1tf} can be viewed as a constrained maximum
likelihood problem with independent observations from a normal
distribution with common variance, $y_t \sim \mbox{N}(\beta_t,
\sigma^2)$, subject to a piecewise linear constraint on
$\beta$. Alternatively, solutions to \eqref{eq:l1tf} are 
maximum a posteriori Bayesian estimators based on Gaussian likelihood
with a special Laplace prior distribution on $\beta$. Note that the
structure of the estimator is determined by the penalty function
$\lambda\norm{D_t\beta}_1$ rather than any parametric trend
assumptions---autoregressive, moving average, sinusoidal seasonal
component, etc. The resulting trend is therefore essentially
nonparametric in the same way that splines are nonparametric. In fact,
using squared $\ell_2$-norm as the penalty instead of $\ell_1$ results exactly in
regression splines.


\subsection{Modifications for Variance}
\label{sec:l1tf_var}

Inspired by the $\ell_1$-trend filtering algorithm, we propose a
non-parametric model for estimating the variance of a time-series. To
this end, we assume that at each time step $t$, there is a parameter
$h_t$ such that the observations $y_t$
are independent normal variables with zero mean and variance
$\exp(h_t)$. The negative log-likelihood of the observed data in this
model is $l(y\given h) \propto -\sum_t h_t - y_t^2e^{-h_t}$. Crucially,
we assume that the parameters $h_t$ vary smoothly and estimate them by
minimizing the penalized, negative
log-likelihood:  
\begin{equation}
  \label{eq:l1tf_var}
  \min_h -l(y\given h)+\lambda \norm{ D_th }_1
\end{equation}
where $D_t$ has the same structure as above.

As with~\eqref{eq:l1tf}, one can solve~\eqref{eq:l1tf_var} using the
PDIP algorithm (as in, e.g.,
\texttt{cvxopt}, \citealp{andersen_cvxopt:_2013}). In each iteration of
PDIP we need to compute a search direction by taking a Newton
step on a system of nonlinear equations. For completeness, we
provide the details in Appendix A of the Supplement, where we show how to
derive the dual of this optimization problem and compute the first and
second derivatives of the dual objective function.

\subsection{Adding Spatial Constraints}
\label{sec:exten}

The method in the previous section can be used to estimate the
variance of a single time-series. Here we extend this
method to the case of spatio-temporal data. 

At a specific time $t$, the data are measured on a grid of points with
$n_r$ rows and $n_c$ columns for a total of $S=n_r\times n_c$ spatial
locations. Let $y_{ijt}$ denote the value of the 
observation at time $t$ on the $i^\text{th}$ row and $j^\text{th}$
column of the grid, and $h_{ijt}$ denote the corresponding
parameter. We seek to impose both temporal and spatial smoothness 
constraints on the parameters. Specifically, we seek a solution
for $h$ which is piecewise linear in time and piecewise constant in
space (although higher-order smoothness can be imposed with minimal
alterations to the methodology). We achieve this goal
by solving the following optimization problem: 
{\small
\begin{align}
  \min_h &\sum_{i,j,t}h_{ijt}+y_{ijt}^2e^{-h_{ijt}}
   +\lambda_t \sum_{i,j} \sum_{t=2}^{T-1}
    \left|h_{ij(t-1)}-2h_{ijt}+h_{ij(t+1)}\right|\label{eq:l1tf_var_st}\\ 
  &+\lambda_s \sum_{t,j} \sum_{i=1}^{n_r-1} \left|h_{ijt}-h_{(i+1)jt}\right|
    +\lambda_s \sum_{t,i} \sum_{j=1}^{n_c-1}
    \left|h_{ijt}-h_{i(j+1)t}\right|\notag
\end{align}
}%

The first term in the objective is proportional to the negative
log-likelihood, the second is the temporal penalty for the
time-series at each location $(i,j)$, while the third and fourth,
penalize the difference between the estimated variance of two
vertically and horizontally adjacent points, respectively. The spatial
component of this
penalty is a special case of trend filtering on
graphs~\cite{WangSharpnack2016} which penalizes the difference between
the estimated values of the signal on the connected nodes (though the
likelihood is different). As before,
we can write \eqref{eq:l1tf_var_st} in matrix form where $h$ is a
vector of length
$TS$ and $D_t$ is replaced by $D \in
\mathbb{R}^{(N_t+N_s) \times (T \cdot S)	}$ (see Appendix C), where $N_t=S \cdot
(T-2)$ and $N_s=T \cdot (2n_rn_c-n_r)$ are the number of temporal and
spatial constraints, respectively. 
Then, as we have two different tuning parameters
for the temporal and spatial components, we write $\Lambda
=\left[\lambda_t\one_{N_t}^\top,\;
  \lambda_s\one_{N_s}^\top\right]^\top$ leading
to:\footnote{Throughout the paper, we use $|x|$ for both scalars and
  vectors. For vectors we use this to denote a vector obtained by
  taking the absolute value of each entry of $x$.}  
\begin{equation}
\min_h -l(y\given h)+ \Lambda^\top | Dh |.
\label{eq:l1tf_var_st_mat}
\end{equation}

\subsection{Related Work}

Variance estimation for financial time series has a lengthy history,
focused especially on parametric models like the generalized
autoregressive conditional heteroskedasticity (GARCH) process~\cite{engle2002dynamic} and
stochastic volatility models~\cite{HarveyRuiz1994}. These models (and
related AR processes) are specifically for parametric modelling of
short ``bursts'' of high volatility, behavior typical of financial
instruments. Parametric models for spatial data go back at least
to~\cite{besag1974spatial} who proposed a conditional probability
model on the lattice for examining plant ecology.

More recently, nonparametric models for both spatial and temporal data
have focused on using $\ell_1$-regularization for trend
estimation. \citet{KimKoh2009} proposed $\ell_1$-trend filtering for
univariate time series, which forms the basis of our methods. These
methods have been generalized to higher order temporal smoothness
~\cite{Tibshirani2014}, graph dependencies~\cite{WangSharpnack2016},
and, most recently, small, time-varying
graphs~\cite{HallacPark2017}. 

Our methodology is similar in flavor to~\cite{HallacPark2017} or
related work in~\citep{GibberdNelson2017,MontiHellyer2014}, but with
several fundamental differences. These papers aim to
discover the time-varying structure of a network. To achieve this
goal, they use Gaussian likelihood with unknown precision matrix and
introduce penalty terms which (1) encourage sparsity among the
off-diagonal elements and (2) discourage changes in the estimated inverse
covariance matrix from one time-step to the next.
Our goal in the present work is to detect the temporal trend in the variance of each
point in the network, but the network is known (corresponding to the grid
over the earth) and fixed in time. To apply these methods in our context
(e.g., \citealp{HallacPark2017}, Eq.\ 2), we would
enforce complete sparsity on the off-diagonal elements (since
they are not estimated) and add a new penalty to enforce spatial
behavior across the diagonal elements. Thus, 
\eqref{eq:l1tf_var_st_mat} is not simply a special case of these
existing methods. Finally, these papers examine networks with hundreds
of nodes and dozens to hundreds of time points. As discussed next, our data are
significantly larger than these networks and attempting to estimate a
full covariance would be prohibitive, were it necessary.

\section{Optimization Methods}
\label{sec:prop-optim-meth}

For a spatial grid of size $S$ and $T$ time steps, $D$ in Equation
\eqref{eq:l1tf_var_st_mat} will have 
$3TS-2S-Tn_r$ rows and $TS$ columns. For a $1^\circ\times
1^\circ$ grid over the entire northern hemisphere and daily data over
10 years, we have $S=90\times 360\approx 32,000$ spatial locations and
$T=3650$ time points, so $D$ has
approximately $10^8$ columns and $10^8$ rows. In principal, we could
solve~\eqref{eq:l1tf_var_st_mat} using PDIP as before, however, each iteration
requires solving a linear system of equations which
depends on $D^\top D$. Therefore,
applying the PDIP directly is infeasible.\footnote{We
  note that $D$ is a highly structured, sparse matrix, but, unlike
  trend filtering alone, it is not banded. We are unaware of general
  linear algebra techniques for inverting such a matrix, despite our
  best efforts to find them.}  

In the next section, we develop two algorithms for solving this
problem efficiently. The first casts the problem as a
so-called consensus optimization problem~\cite{boyd_distributed_2011}
which solves smaller sub-problems using PDIP and then recombines the results. The
second uses proximal methods to avoid matrix inversions. Either may be more
appropriate depending on the particular computing infrastructure.

\subsection{Consensus Optimization}
\label{sec:consOpt}

Consider an optimization problem of the form $\min_h f(h)$, where
$h\in\mathbb{R}^n$ is the global variable and
$f(h):\mathbb{R}^n \rightarrow \mathbb{R}\cup \{+\infty\}$ is
convex. Consensus optimization breaks this problem
into several smaller sub-problems that can be solved independently in
each iteration.  

Assume it is possible to define a set of local variables
$x_i \in \mathbb{R}^{n_i}$ such that $f(h)=\sum_i f_i(x_i)$, where
each $x_i$ is a subset of the global variable $h$. More specifically,
each entry of the local variables corresponds to an entry of the
global variable. Therefore we can define a mapping $G(i,j)$
from the local variable indices into the global variable indices:
$k=G(i,j)$ means that the $j^\text{th}$ entry of $x_i$ is
$h_k$ (or $(x_i)_j=h_k$). For ease of notation, define $\tilde{h}_i
\in \mathbb{R}^{n_i}$ as $(\tilde{h}_i)_j=h_{G(i,j)}$. Then,
the original optimization problem is equivalent to:  
\begin{equation}
\begin{aligned}
&\min_{\{x_1,...,x_N  \}} \sum_{i=1}^N f_i(x_i) &
s.t. &\quad \tilde{h}_i=x_i.
\end{aligned}
\label{eq:consADMM}
\end{equation}
It is important to note that each entry of the global variable may
correspond to several entries of the local variables and so the
constraints $\tilde{h}_i=x_i$ enforce consensus between the local
variables corresponding to the same global variable.  
The augmented Lagrangian corresponding to 
\eqref{eq:consADMM} is $L_\rho(x,h,y)=\sum_i
\big(f_i(x_i)+u_i^\top(x_i-\tilde{h}_i) + (\rho/2) \lVert
x_i-\tilde{h}_i \lVert_2^2 \big)$. Now, we can apply ADMM to
$L_\rho$. This results in solving $N$ independent optimization
problems followed by a step to achieve consensus among the solutions
in each iteration. 
To solve the optimization problem \eqref{eq:l1tf_var_st_mat} using
this method, we need to address two 
questions: first, how to choose the 
local variables $x_i$, and second, how to the update them.
\begin{figure}[tb]
  \centering
  \includegraphics[width=.4\textwidth]{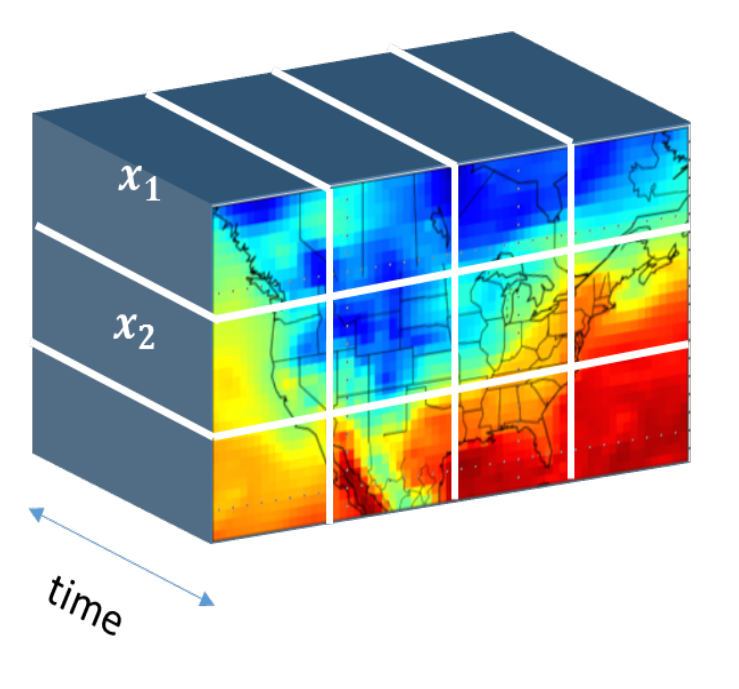}
  \caption{The cube represents the global variable $h$ in space and
    time. The sub-cubes specified by the white lines are
    $x_i$.}
  \label{fig:data_cube}
\end{figure}

In \autoref{fig:data_cube}, the global variable $h$ is represented
as a cube. We decompose $h$
into sub-cubes as shown by white lines. Each global variable inside
the sub-cubes corresponds to only one local variable. The global
variables on the border (white lines), however, correspond to more
than one local variable. With 
this definition of $x_i$, the objective
\eqref{eq:l1tf_var_st_mat} decomposes as $\sum_i f_i(x_i)$ where
$f_i(x_i)=-l(y_i\given x_i)+\Lambda_{(i)}^\top |D_{(i)}x_i|$, and
$\Lambda_{(i)}$ and $D_{(i)}$ contain the temporal and spatial
penalties corresponding to $x_i$ only in one sub-cube along with its
boundary. Thus, we now need to use PDIP to solve $N$ problems each of
size $n_i$, which is feasible for small enough
$n_i$. \autoref{alg:conADMM} gives the general version of this
procedure. A more detailed discussion of this is in Appendix B of the
Supplement where we show how to compute the dual and the derivatives
of the augmented Lagrangian.  

\begin{algorithm}[tb]
  \caption{Consensus ADMM }
  \label{alg:conADMM}
  \begin{algorithmic}[1]
    \STATE {\bfseries Input:} data $y$, penalty matrix $D$, 
    $\epsilon, \rho,\lambda_t,\lambda_s >0$.
    \STATE {\bf Set:} $h\leftarrow 0$, $z\leftarrow 0$, $u\leftarrow
    0$. \COMMENT{Initialization} 
    \REPEAT
    \STATE $\begin{aligned}x_i&\leftarrow\argmin_{x_i} -l(y_i\given
    x_i)+\Lambda_{(i)}^\top |D_{(i)}x_i|+ (u_i)^\top x_i +
    (\rho/2)  \lVert x_i-\tilde{h}_i \rVert_2^2\end{aligned}$ \COMMENT{Update local
      vars using PDIP}
    \STATE $h_k\leftarrow (1/S_k)\sum_{G(i,j)=k} (x_i)_j
    $. \COMMENT{Global update.}
    \STATE $ u_i\leftarrow u_i + \rho (x_i-\tilde{h}_i)$. \COMMENT{Dual update}
    \UNTIL {$\max\left\{\norm{h^{m+1}-h^m},\ \norm{h^m-x^m}\right\} < \epsilon$}
    \STATE {\bf Return:} $h$.
  \end{algorithmic}
\end{algorithm}

Because consensus ADMM breaks the large optimization into
sub-problems that can be solved independently, it is amenable to a
split-gather parallelization strategy via, e.g., the MapReduce framework.
In each iteration, the
computation time will be equal to the time to solve each sub-problem
plus the time to communicate the solutions to the master processor
and perform the consensus step. Since each sub-problem is
small, with parallelization, the computation time in each iteration
will be small. In addition, our experiments with several values of
$\lambda_t$ and $\lambda_s$ showed that the algorithm converges in a few
hundred iterations. 
This algorithm is most useful if we can parallelize the
computation over several machines with low communication cost between
machines. In the next section, we describe 
another algorithm which makes the computation feasible on a single
machine. 

\subsection{Linearized ADMM}
\label{sec:linADMM}


Consider the generic optimization problem
$\min_x f(x)+g(Dx)$
where $x\in \mathbb{R}^n$ and $D\in \mathbb{R}^{m\times n}$. Each
iteration of the linearized ADMM
algorithm~\cite{parikh_proximal_2014} for solving this problem 
has the form
\begin{equation*}
  \label{eq:linADMM_steps}
  \begin{aligned}
    x & \leftarrow \prox_{\mu f} \left(x - (\mu/\rho)D^\top (D x - z + u )\right)\\
    z & \leftarrow \prox_{\rho g} \left(Dx + u\right)\\
    u & \leftarrow u + D x - z
  \end{aligned}
\end{equation*}
where the algorithm parameters $\mu$ and $\rho$ satisfy $0 < \mu <
\rho/\norm{D}_2^2$, $z,u\in \mathbb{R}^m$ and the proximal operator is
defined as $\prox_{\alpha \varphi}(u) = \min_x \,\, \alpha \cdot
\varphi(x)+\frac{1}{2} \norm{ x-u}_2^2$. Proximal algorithms are feasible
when these proximal operators can be evaluated efficiently which, as
we show next, is the case.  


\begin{lemma}
  Let $f(x) = \sum_{k} x_{k} + y_{k}^2e^{-x_{k}}$ and $g(x) =
  \norm{x}_1$. Then, 

  \begin{equation*}
    \begin{aligned}
      \bigl[\prox_{\mu f}(u) \bigl]_k &= \mathscr{W}\bigg(\frac{y_k^2}{\mu}
      \exp\bigg(\frac{1-\mu u_k}{\mu}\bigg) \bigg) + \frac{1-\mu u_k}{\mu},\\
      \prox_{\rho g}(u) &= S_{\rho\lambda}(u)
    \end{aligned}
  \end{equation*}
  where $\mathscr{W}(\cdot)$ is the Lambert W function
  \cite{corless_lambertw_1996},  $[S_{\alpha}(u)]_k = \sign(u_k)(|u_k|
  -\alpha_k)_+$ and $(v)_+=v\vee 0$.
\end{lemma}
The proof is fairly straightforward and given
in Appendix C in the Supplement.  
Therefore, \autoref{alg:linADMM} gives a different method for solving
the same problem. In this case, both the primal update and the
soft thresholding step are performed elementwise at each point of the
spatio-temporal grid. It can therefore be extremely fast to perform
these steps. However, because there are now many more dual variables,
this algorithm will require more outer iterations to achieve
consensus. It therefore is highly problem and architecture dependent
whether \autoref{alg:conADMM} or \autoref{alg:linADMM} will be more
useful in any particular context. In our experience,
\autoref{alg:conADMM} requires an order of magnitude fewer iterations, but
each iteration is much slower unless carefully parallelized.

\begin{algorithm}[tb]
  \caption{Linearized ADMM }
  \label{alg:linADMM}
  \begin{algorithmic}[1]
    \STATE {\bfseries Input:} data $y$, penalty matrix $D$, 
    $\epsilon, \rho,\lambda_t,\lambda_s >0$.
    \STATE {\bf Set:} $h\leftarrow 0$, $z\leftarrow 0$, $u\leftarrow
    0$. \COMMENT{Initialization} 
    \REPEAT
    \STATE $h_k\leftarrow \mathscr{W}\bigg(\frac{y_k^2}{\mu} 
    \exp\bigg(\frac{1-\mu u_k}{\mu}\bigg) \bigg) + \frac{1-\mu
      u_k}{\mu}$ for all $k=1,\ldots TS$. \COMMENT{Primal update}
    \STATE $z\leftarrow S_{\rho\lambda}(u)$. \COMMENT{Elementwise soft thresholding}
    \STATE $u\leftarrow u - z$. \COMMENT{Dual update}
    \UNTIL {$\max\{\norm{Dh-z},\; \norm{z^{m+1}-z^m}\} < \epsilon$}
    \STATE {\bf Return:} $z$.
  \end{algorithmic}
\end{algorithm}

\section{Empirical Evaluation}
\label{sec:empirical-evaluation}

In this section, we examine both simulated and real spatio-temporal
climate data. All the computations were performed on a Linux machine
with four 3.20GHz Intel i5-3470 cores. 

\subsection{Simulations}
\label{sec:simulations}

\begin{figure}[tb]
  \centering	 \raisebox{.03\textheight}{\includegraphics[height=.1\textheight]{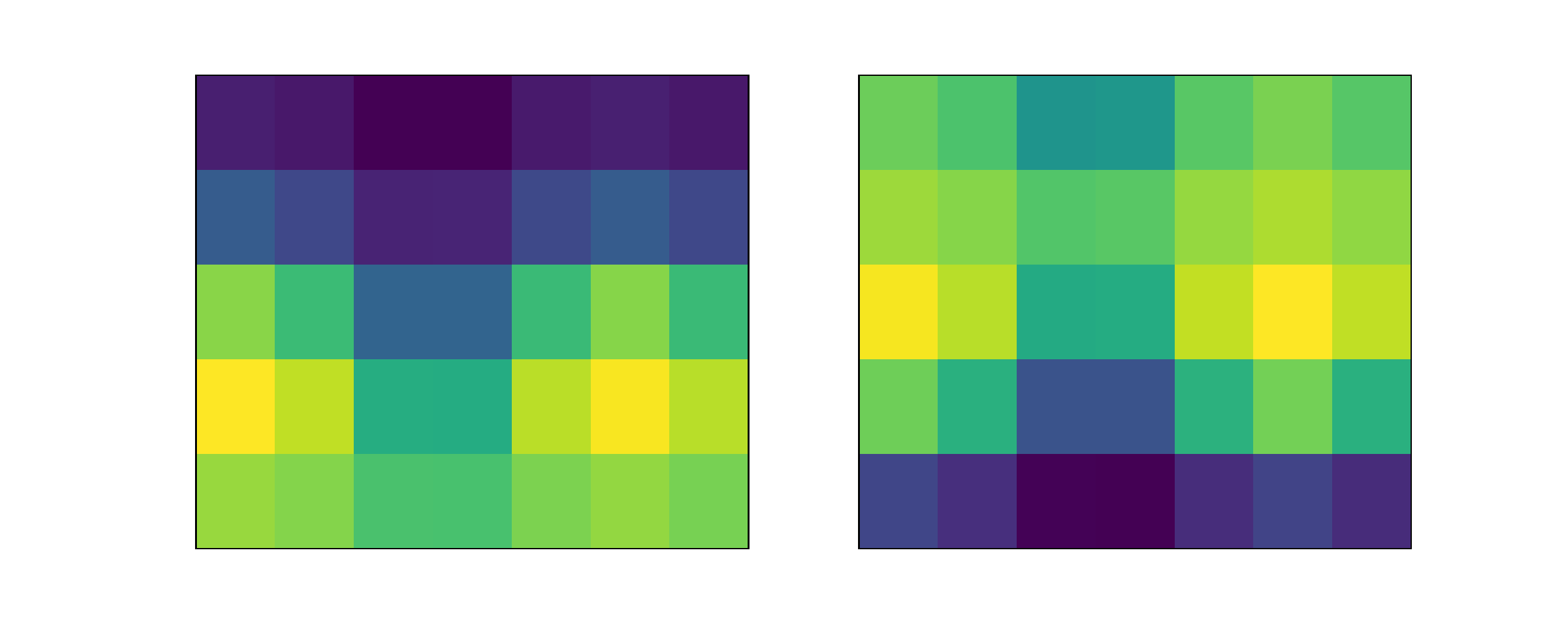}}
  \includegraphics[height=.15\textheight]{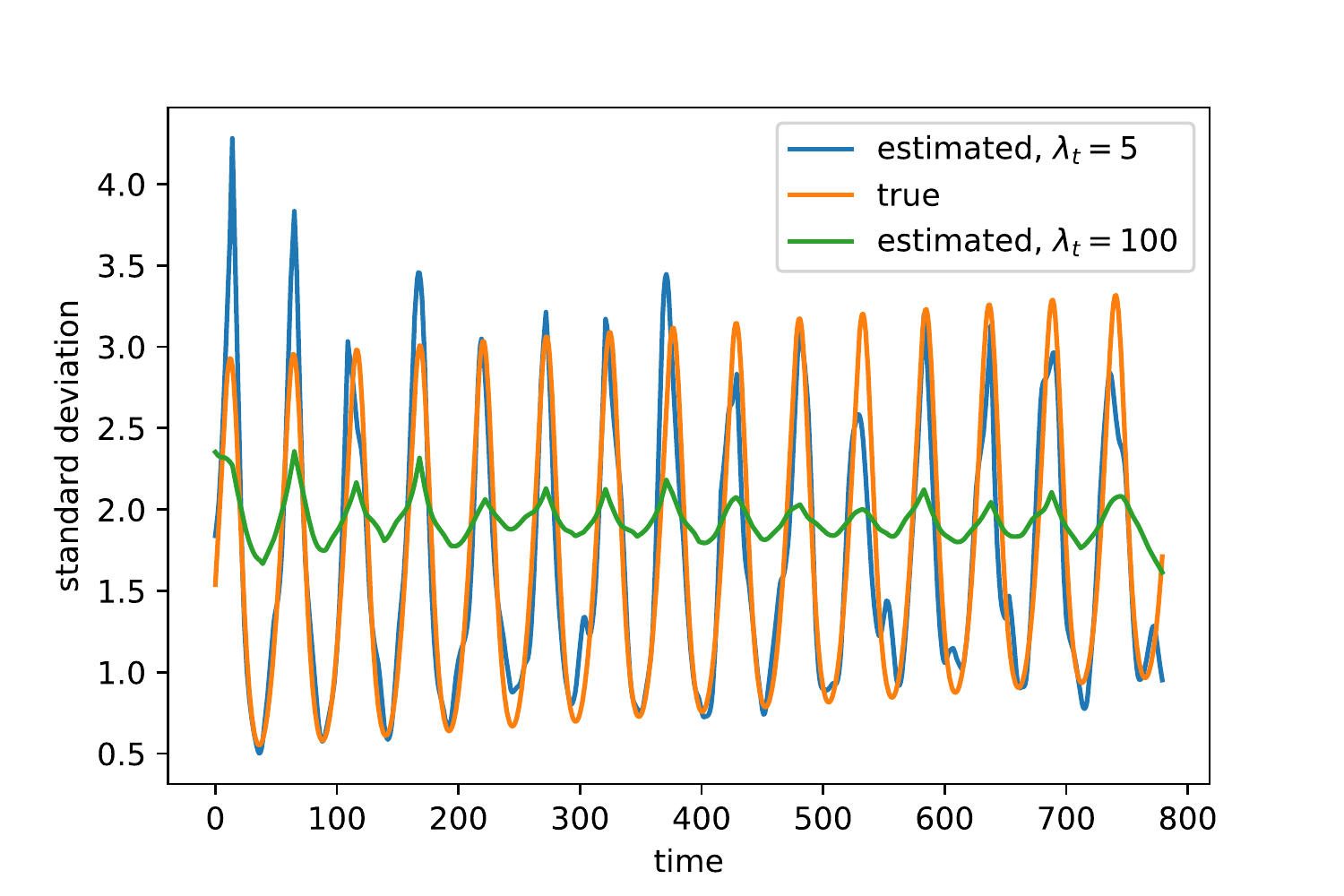}
  \caption{Left: Variance function at $t=25$ (left) and $t=45$
    (right). Right: The true (orange) and estimated standard deviation 
    function at the location (0,0). The estimated values are
    obtained using linearized ADMM with $\lambda_s=0.1$ and two
    values of $\lambda_t$: $\lambda_t=5$ (blue) and
    $\lambda_t=100$ (green).} \label{fig:true_var_spatial}
\end{figure}

Before examining real data, we apply our model to some synthetic
data. This example was constructed to mimic the types of spatial and
temporal phenomena observable in typical climate data. We generate a
complete spatio-temporal field wherein 
observations at all time steps and all locations are
independent Gaussian random variables with zero mean. However, the
variance of these random variables follows a smoothly varying function
in time and space given by the following parametric model:
\begin{align*}
  \sigma^2(t,r,c) & =\sum_{k=1}^{K} W_k(t) \cdot \exp\bigg(
                    \frac{(r-r_k)^2+(c-c_k)^2} {2\sigma_k^2} \bigg)\\
  W_k(t) & =\alpha_k \cdot t + \exp(\sin(2\pi\omega_k t+\phi_k)) .
\label{eq:sourceVar}
\end{align*}
The variance at each time and location is computed as the
weighted sum of $K$ bell-shaped functions where the weights are
time-varying, consist of a linear trend 
and a periodic term. 
The
bell-shaped functions impose spatial smoothness while the linear
trend and the periodic terms enforce the temporal smoothness similar
to the seasonal component in real climate data. We simulated the
data on a 5$\times$7 grid for 780 time steps with $K=4$. This yields a
small enough problem to be evaluated many times while
still mimicking important properties of climate data.
Specific parameter choices of the variance function are shown in
\autoref{tab:sim_params}.
For illustration, we also plot the
variance function for all locations at $t=25$ and $t=45$
(\autoref{fig:true_var_spatial}, top panel) as well as
the variance across time at location $(0,0)$ (\autoref{fig:true_var_spatial}, bottom panel, orange).
\begin{table}[tb]
  \caption{Parameters used to simulate data.}
  \label{tab:sim_params}
  \begin{center}
    \begin{tabular}{@{}lllllll@{}}
      \toprule
      $s$ & $r_s$ & $c_s$ & $\sigma_s$ &$\alpha_s$ & $\omega_s$ & $\phi_s$\\
      \midrule
      1 & 0 & 0 & 5 & 0.5 & 0.121 & 0 \\
      2 & 0 & 5 & 5 & 0.1 & 0.121 & 0 \\
      3 & 3 & 0 & 5 & -0.5 & 0.121 & $\pi/2$ \\
      4 & 3 & 5 & 5 & -0.1 & 0.121 & $\pi/2$ \\
      \bottomrule
    \end{tabular}
  \end{center}
\end{table} 

We estimated the linearized ADMM for all combinations of values of
$\lambda_t$ and $\lambda_s$ from the sets $\lambda_t \in
\{0,1,5,10,50,100\}$ and $\lambda_s \in \{0,0.05,0.1,0.2,0.3\}$. For
each pair, we then compute the mean absolute error (MAE) between the
estimated variance and the true variance at all locations and all time
steps. For $\lambda_t=5$ and $\lambda_s=0.1$, the MAE was minimized. 
\begin{figure}[tb]
  \centering	
  \includegraphics[width=.48\columnwidth]{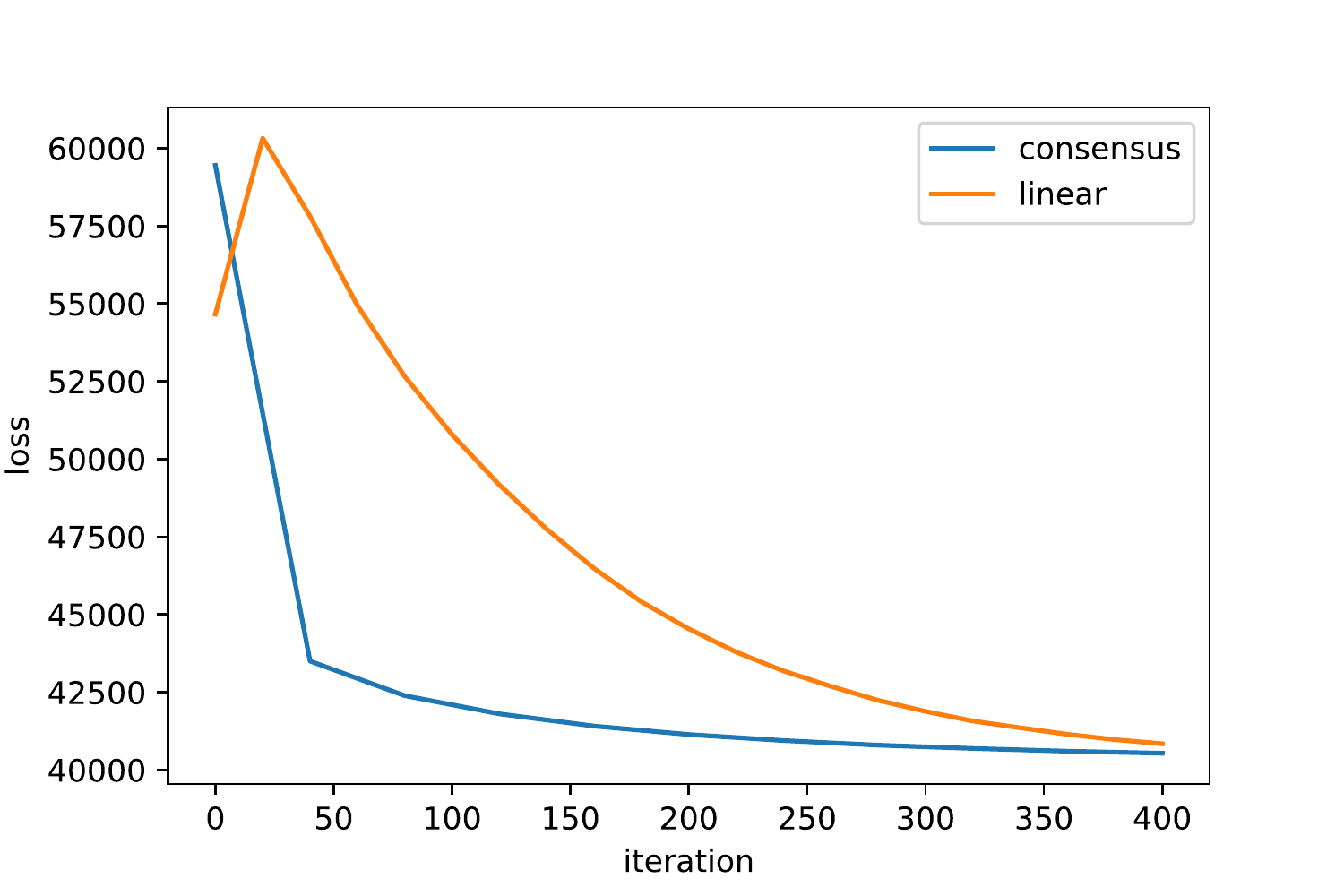}  
  \includegraphics[width=.48\columnwidth]{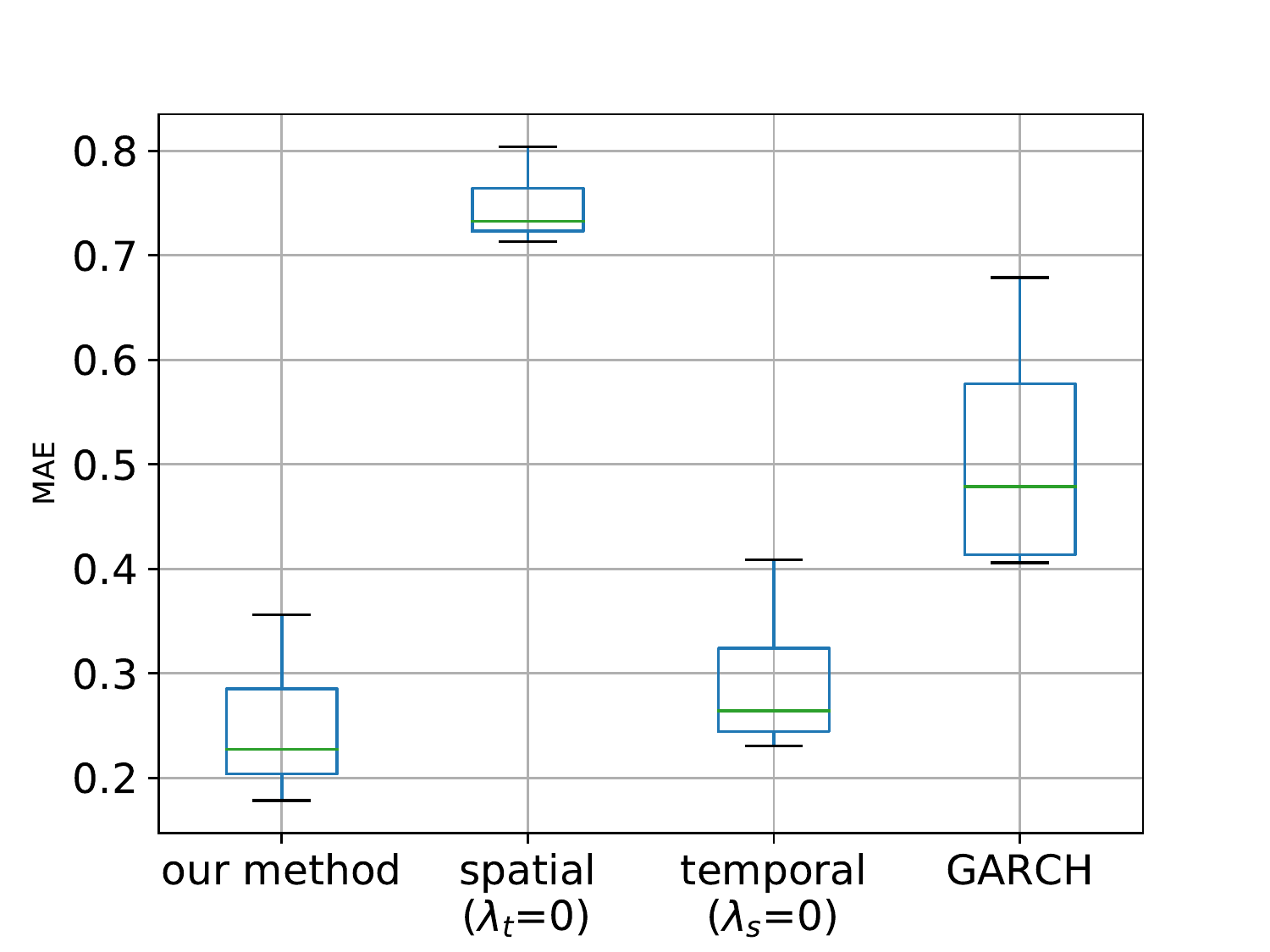}  
  \caption{Left: Value of the objective function for linearized (orange)
    and consensus (blue) ADMM against iteration. Right: MAE for (1) our
    method with optimal values of $\lambda_t$ and $\lambda_s$ (2)
    spatial penalty only (3) temporal
    penalty only and (4) a GARCH(1,1).} 
  \label{fig:true_fitted_var}
\end{figure}
The
bottom panel of \autoref{fig:true_var_spatial} shows the true and the
estimated standard deviation at location (0,0) 
and $\lambda_t=5$ (blue) and $\lambda_t=100$ (green) ($\lambda_s=0.1$).
Larger values of $\lambda_t$ lead to estimated values
which are ``too smooth". The left panel of \autoref{fig:true_fitted_var} shows the
convergence of both Algorithms as a function of iteration. It is
important to note that each iteration of the linearized
algorithm takes 0.01 seconds on average while each iteration of the
consensus ADMM takes about 20 seconds. Thus, where the lines meet at
400 iterations requires about 4 seconds for the linearized method and
2 hours for the consensus method. For consensus ADMM, computation per
iteration per core requires $\sim$10 seconds with $\sim$4 seconds
for communication. In general, the literature suggests linear
convergence for ADMM~\citep{NishiharaLessard2015} and~\autoref{fig:true_fitted_var} seems to fit in the linear framework for both algorithms, though with different constants.

To further examine the performance of the proposed model, we next
compare it to three alternatives: a model which does not consider the
spatial smoothness (equivalent to fitting the model in
\autoref{sec:l1tf_var} to each time-series separately), a model which only imposes spatial smoothness, and a GARCH(1,1) model. We
simulated 100 datasets using the method explained above with $\sigma_s
\sim \mathsf{uniform}(4,7)$. The right panel of
\autoref{fig:true_fitted_var} shows the boxplot of the MAE for these
models. As discussed above, using an algorithm akin
to~\citep{HallacPark2017} ignores the spatial component and thus gives
results which are similar to the second column if the covariances are shrunk to
zero (massively worse if they are estimated).

\subsection{Data Analysis}
\label{sec:data}

Consensus ADMM in~\autoref{sec:consOpt} is appropriate when we can
easily parallelize over multiple machines.  Otherwise, it is significantly
slower, so all the results reported in this section are obtained
using \autoref{alg:linADMM}. We
applied this algorithm to the Northern Hemisphere of the ERA-20C
dataset available from the European Center for Medium-Range
Weather Forecasts\footnote{\url{https://www.ecmwf.int}}. We use 
the 2 meter temperature measured daily at noon local time from January 1, 1960
to December 24, 2010. 

\paragraph{Preprocessing and other considerations}

Examination of the time-series alone demonstrates
strong differences 
between trend and cyclic behavior across spatial locations (the data
are not mean-zero). One might
try to model the cycles by the summation of 
sinusoidal terms with different frequencies. However, for some
locations, this would require many frequencies
to achieve a reasonable level of accuracy, while other locations would
require relatively few. In addition, such a model
cannot capture the non-stationarity in the cycles. 

\begin{figure}[tb]
	\centering
	\includegraphics[width=.66\columnwidth]{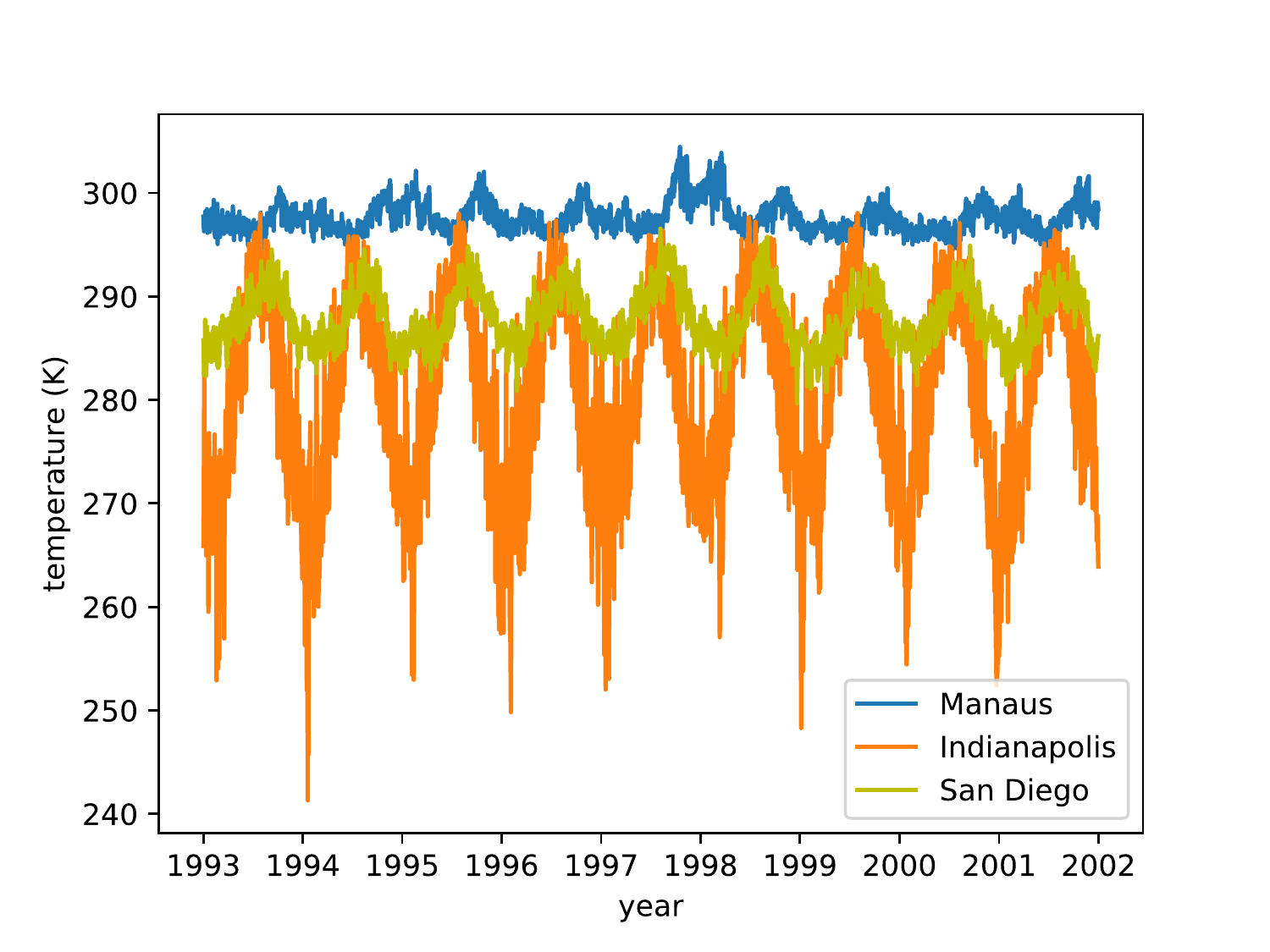}
 	\caption{Time-series of the temperature (in Kelvin) of three cities.}
 	\label{fig:cities_ts}
\end{figure}

\autoref{fig:cities_ts} shows the time-series of the 
temperature of three cities: Indianapolis (USA), San Diego (USA) and
Manaus (Brazil). The time-series of Indianapolis and San Diego show
clear cyclic behavior, though the amplitude is different. The
time-series of Manaus does not 
show any regular cyclic behavior. For this reason, we
first apply trend filtering to remove seasonal terms and detrend
every time-series. For each time-series, we found the optimal value of
the penalty parameter using 5-fold cross-validation.

\begin{figure}[tb]
  \centering
  \includegraphics[width=.8 \columnwidth]{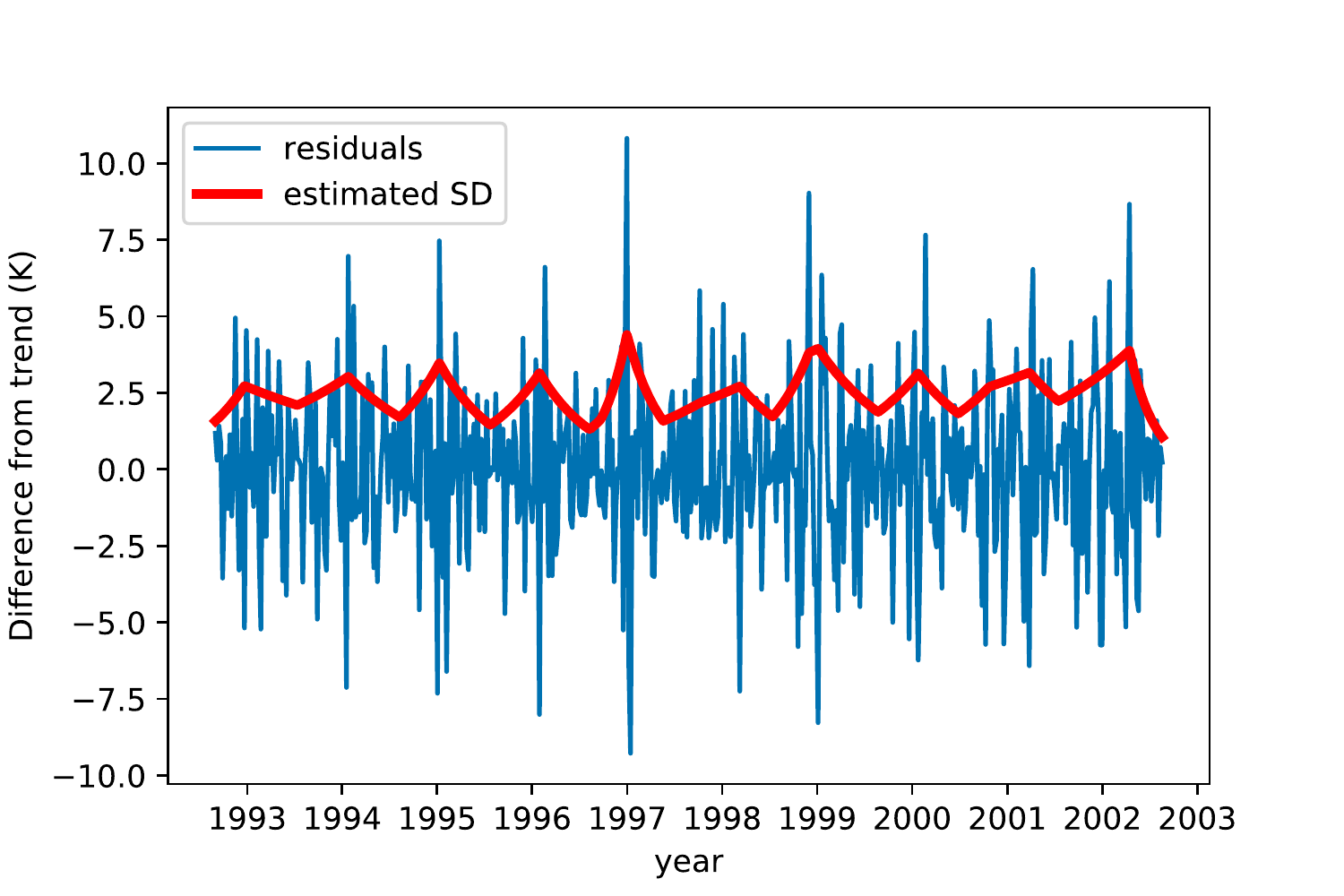}
  \includegraphics[width=.45 \columnwidth]{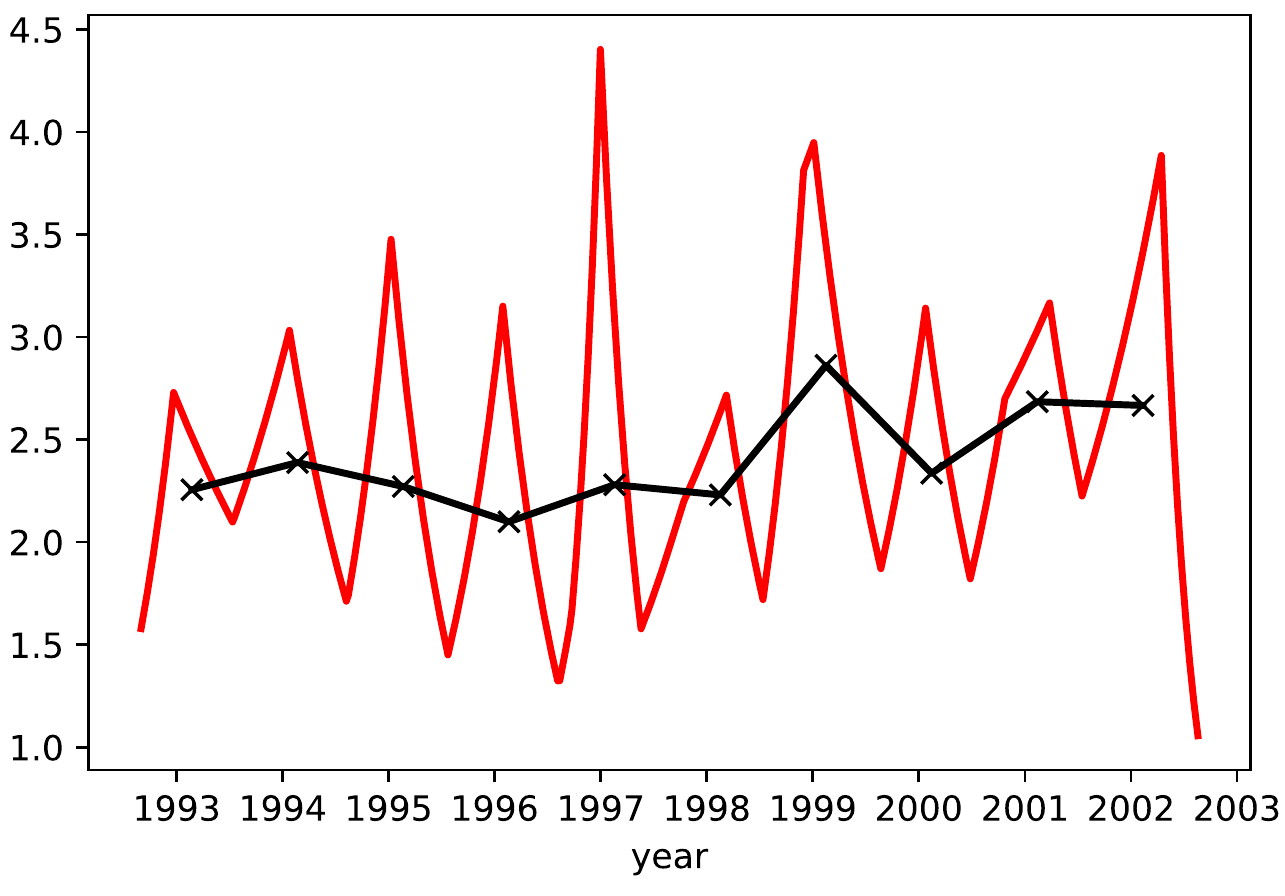}
  \includegraphics[width=.45 \columnwidth]{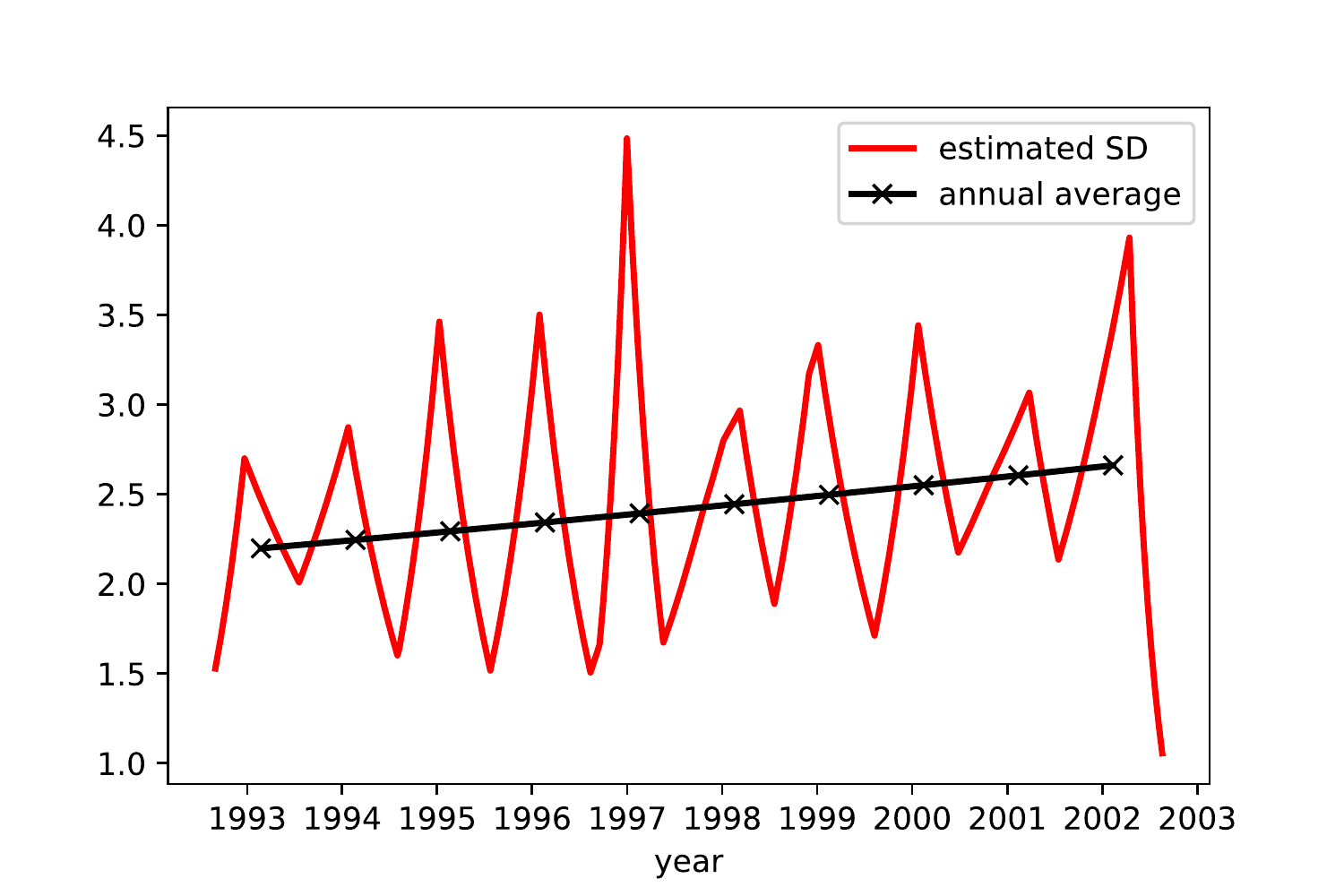}  
  \caption{Top: The variability of the time-series of Indianapolis
    (weekly) and the estimated SD obtained from the method of
    \autoref{sec:l1tf_var} (red). Lower left: the estimated SDs (red) and
    their annual average (black) without the long horizon
    penalty. Lower right: the same but with the long horizon
    penalty.} 
  \label{fig:bloom_estimatedSD}
\end{figure}

The blue curve in the top panel of 
\autoref{fig:bloom_estimatedSD} shows the daily
temperature for Indianapolis after detrending. 
The red curve 
shows the estimated SD, $\exp(h_t/2)$, obtained from our
proposed model. For ease of analysis, we compute the average
of the estimated SD for each year. Both are shown in the lower left panel of
\autoref{fig:bloom_estimatedSD}.

In addition to the constraints discussed in \autoref{sec:exten}, we add
a long horizon penalty to smooth the annual trend:
$\sum_{i=2}^{N_{\y}-1}\left| \sum_t h_{\mathcal{A}(-1)}-2h_{\mathcal{A}(0)}+h_{\mathcal{A}(1)}\right|$
where $N_{\y}$ is the number of years and $\mathcal{A}(b)=\{t : t\in
(\y_i+b)\}$. Finally, because the observations are on the
surface 
of a hemisphere rather than a grid, we add extra spatial constraints with the
obvious form to handle the boundary between
$180^{\circ}$W and $180^{\circ}$E as well as the region near the North Pole.
The estimated SDs for Indianapolis are shown in the lower right panel of \autoref{fig:bloom_estimatedSD}. The annual average
of the estimated SDs shows a linear trend with a positive slope. 




As shown in \autoref{alg:linADMM}, we checked convergence using
$\epsilon=0.001\%$ of the MSE of the data.
Our simulations indicated that the
convergence speed depends on the value of $\lambda_t$ and
$\lambda_s$. For the temperature data, we used the solutions obtained for smaller values
of these parameters as warm starts for larger values. Estimation takes
between 1 and 4 hours for convergence for each pair of tuning parameters.

\paragraph{Model Selection}
One common method for choosing the penalty parameters in lasso
problems is to find the solution for a range of the values of these
parameters and then choose the values which minimize a model selection
criterion. However, such analysis needs either the computation of the degrees
of freedom or requires cross validation. Previous work has
investigated the degrees of freedom in 
generalized lasso problems with Gaussian likelihood
\cite{tibshirani_degrees_2012,hu_dual_2015,zeng_geometry_2017}, but,
results for non-Gaussian likelihood remains an open problem, and cross
validation is too expensive. 
In this paper, therefore, we use a heuristic method for choosing $\lambda_t$ and
$\lambda_s$: we compute the solutions for a range of values of
and choose those which minimize
$L(\lambda_t,\lambda_s)=-l(y|h)+ \sum \lVert D h
\lVert$. This objective is a compromise between the negative log
likelihood and the complexity of the solution. For smoother solutions the value of $\sum
\lVert Dh \lVert$ will be smaller but with the cost of larger
$-l(y|h)$. We computed the solution for all the combinations of the
following sets of values: $\lambda_t \in \{0,2,4,8,10,15,200,1000\} \, \, ,
\lambda_s \in \{0,.1,.5,2,5,10\}$. The best combination was
$\lambda_t=4$ and $\lambda_s=2$.

\paragraph{Analysis of Trends in Temperature Volatility}

\begin{figure}[tb]
  \centering
  \includegraphics[width=.45\columnwidth]{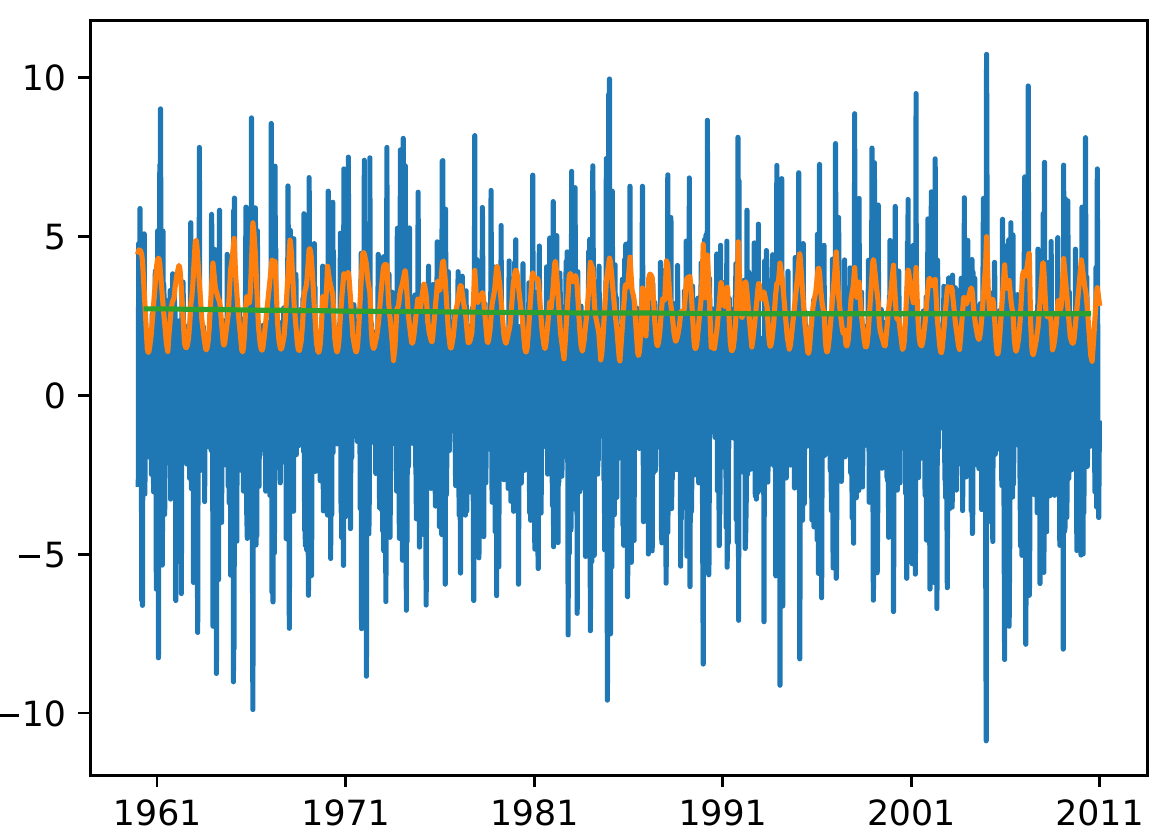}
  \includegraphics[width=.45\columnwidth]{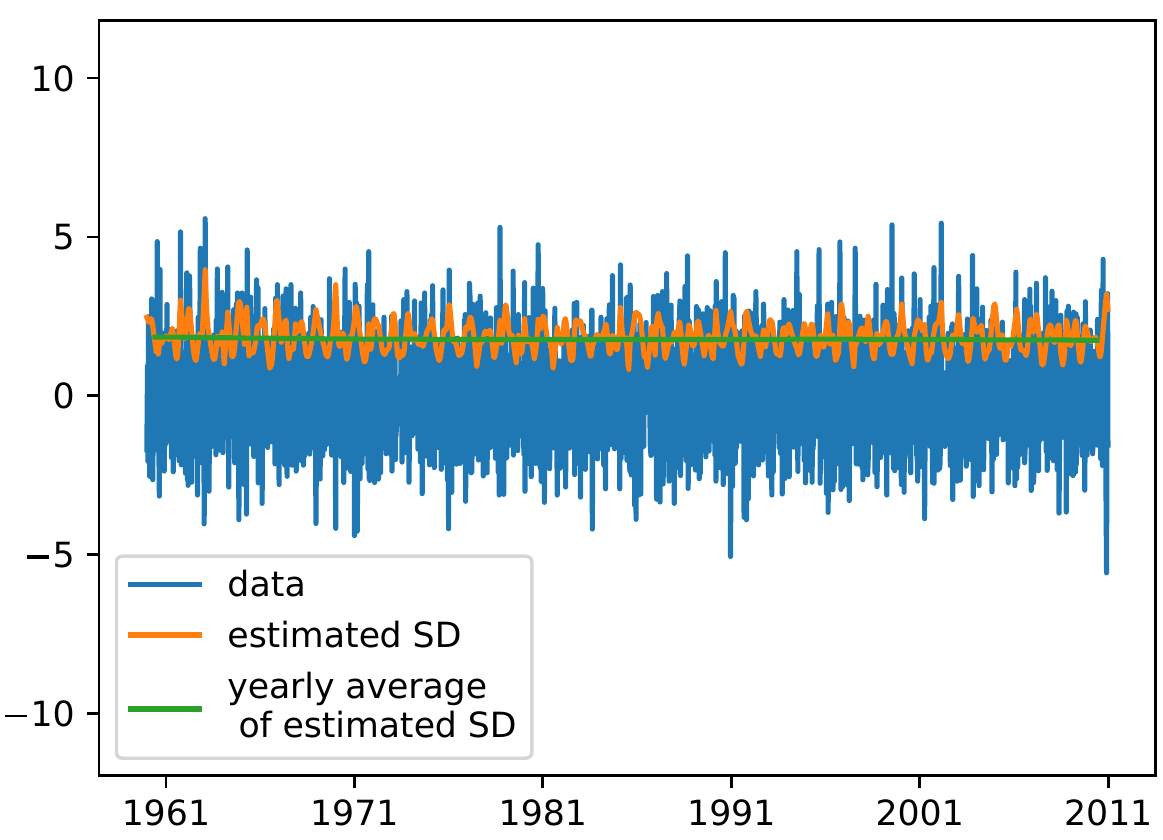}
  \caption{Residuals from the estimated trend
    (blue), the estimated SDs (orange), and annual average SD (green) for
   Indianapolis (left) and San Diego (right). Units are $\textrm{K}^{\circ}$.} 
  \label{fig:avg_change_estimatedSD}
\end{figure} 

\autoref{fig:avg_change_estimatedSD} shows the
detrended data, the estimated standard deviation and the yearly
average of these estimates for two cities in the US: Indianapolis
(left) and San Diego (right). The estimated SD 
captures the periodic behavior in the variance of the time-series. In
addition, the number of linear segments changes adaptively in each
time window depending on how fast the variance is changing.  
\begin{figure}[tb]
  \centering
  \includegraphics[width=.9\linewidth]{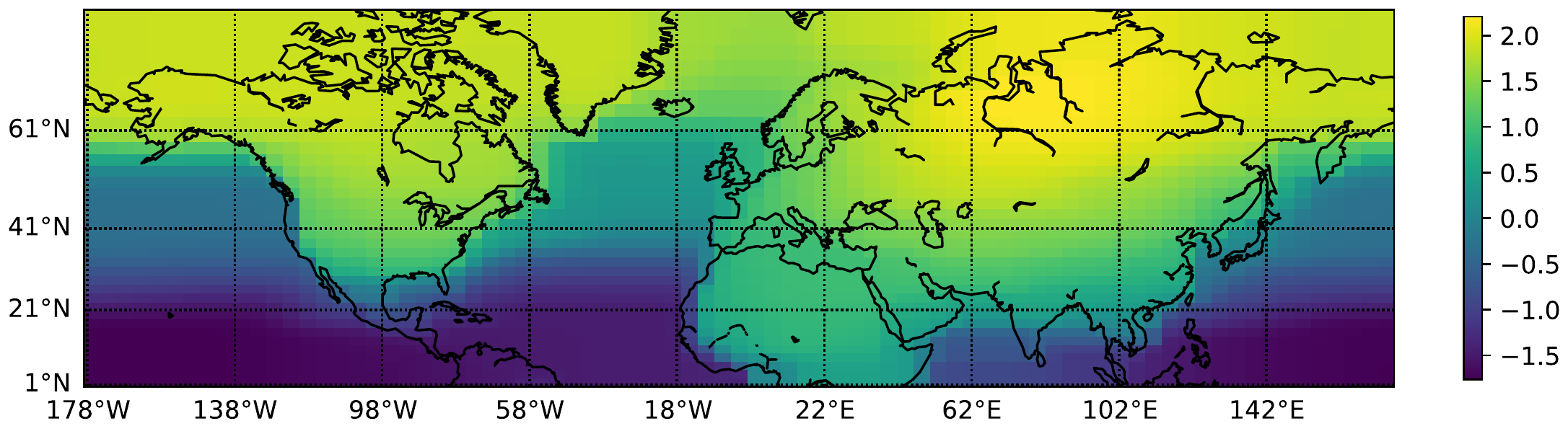}\\
  \includegraphics[width=.9\linewidth]{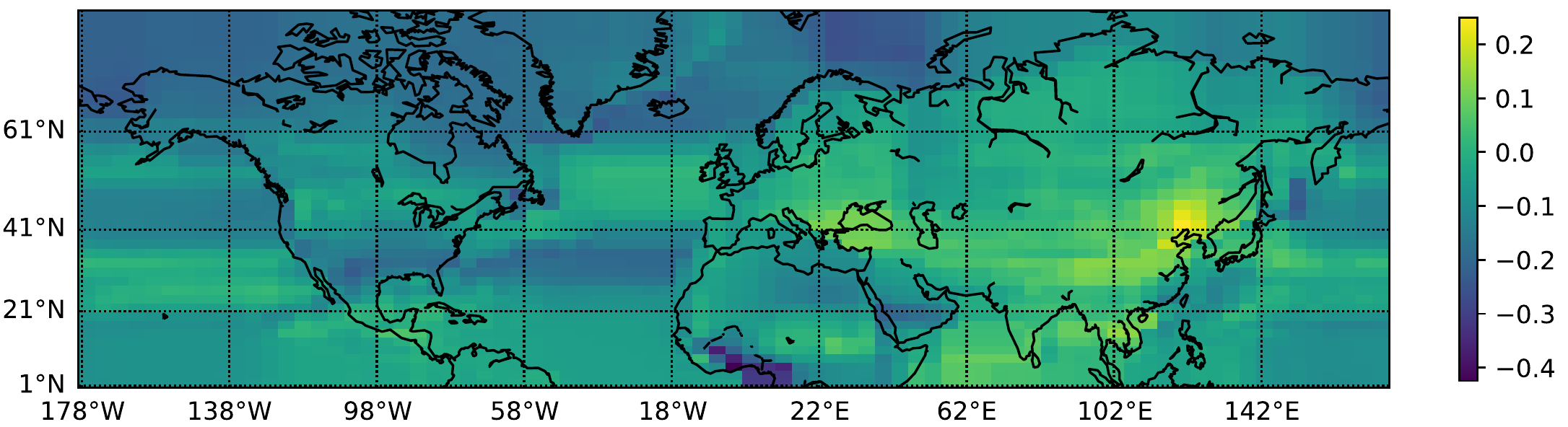}
  \caption{The
    average of the detrended estimated variance over the northern hemisphere (top) and the
    change in the variance from 1961 to 2011 (bottom). Units are K$^\circ$.}
\label{fig:avg_chg}
\end{figure}

The yearly average of the estimated SD captures the trend in the
temperature volatility. For example, we can see that the variance in
Indianapolis displays a small positive trend (easiest to see in
\autoref{fig:bloom_estimatedSD}). To determine how the volatility has 
changed in each location, we subtract the average of the estimated
variance in 1961 from the average in the following years and compute
their sum. The average estimated variance at each location is shown in
the top panel of~\autoref{fig:avg_chg} while the change from 1961
is depicted in bottom panel. Since the
optimal value of the spatial penalty is rather large ($\lambda_s=2$)
the estimated variance is spatially very smooth. 

The SD in most locations on the northern
hemisphere had a negative trend in this time period, though spatially,
this decreasing pattern is localized mainly toward the extreme
northern latitudes and over oceans. In many ways, this is consistent
with climate change predictions: oceans tend to operate as a local
thermostat, regulating deviations in local temperature, while warming polar
regions display fewer days of extreme cold.
The most positive trend can be observed in Asia,
particularly South-East Asia. 



\section{Discussion}
\label{sec:discussion}

In this paper, we proposed a new method for estimating the variance of
spatio-temporal data with the goal of analyzing global
temperatures. The main idea is to cast this problem as a 
constrained optimization problem where the constraints enforce smooth
changes in the variance for neighboring points in time and space. In
particular, the solution is piecewise linear in time and piecewise
constant in space. The resulting optimization is in the form of a
generalized LASSO problem with high-dimension, and so applying the
PDIP method directly is infeasible. We therefore developed two
ADMM-based algorithms to solve this problem: the consensus ADMM and
linearized ADMM. 

The consensus ADMM algorithm converges in a few hundred iterations
but each iteration takes much longer than the linearized ADMM
algorithm. The appealing feature of the consensus ADMM algorithm is
that if it is parallelized on enough machines the
computation time per iteration remains constant as the problem size
increases. The linearized ADMM algorithm on the other hand converges
in a few thousand iterations but each iteration is performed in a
split second. However, since the algorithm converges in many
iterations it is not very appropriate for parallelization. The reason
is that after each iteration the solution computed on each local machine
must be collected by the master machine, and this operation takes
depends on the speed of the network connecting the
slave machines to the master. A direction for future research would be
to combine these two algorithms in the following way: the problem
should be split into the sub-problems (as in the consensus ADMM) but
each sub-problem can be solved using linearized ADMM. 



\section*{Acknowledgements}

This material is based upon work supported by the National Science
Foundation under Grant Nos.\ DMS--1407439 and DMS--1753171.

\fontsize{9.0pt}{10.0pt} \selectfont
\bibliographystyle{mybibsty}
\bibliography{aaai-references}

\begin{thebibliography}{999}
\newcommand{\enquote}[1]{``#1''}

\bibitem[Andersen et~al.(2013)Andersen, Dahl, and
  Vandenberghe]{andersen_cvxopt:_2013}
{\sc Andersen, M.~S., Dahl, J., and Vandenberghe, L.} (2013), \emph{{CVXOPT}:
  {A} {Python} package for convex optimization, version 1.1.6.}, Available at
  \texttt{cvxopt.org}.

\bibitem[Bender et~al.(2012)Bender, Ramanathan, and
  Tselioudis]{BenderRamanathan2012}
{\sc Bender, F.~A., Ramanathan, V., and Tselioudis, G.} (2012),
  \enquote{Changes in extratropical storm track cloudiness 1983--2008:
  Observational support for a poleward shift,} \emph{Climate Dynamics}, {\bf
  38}(9-10), 2037--2053.

\bibitem[Besag(1974)Besag]{besag1974spatial}
{\sc Besag, J.} (1974), \enquote{Spatial interaction and the statistical
  analysis of lattice systems,} \emph{Journal of the Royal Statistical Society.
  Series B (Methodological)}, {\bf 36}, 192--236.

\bibitem[Bony et~al.(2015)Bony, Stevens, Frierson, Jakob, Kageyama, Pincus,
  Shepherd, Sherwood, Siebesma, Sobel, et~al.]{BonyStevens2015}
{\sc Bony, S., Stevens, B., Frierson, D.~M., Jakob, C., Kageyama, M., Pincus,
  R., Shepherd, T.~G., Sherwood, S.~C., Siebesma, A.~P., Sobel, A.~H., et~al.}
  (2015), \enquote{Clouds, circulation and climate sensitivity,} \emph{Nature
  Geoscience}, {\bf 8}(4), 261.

\bibitem[Boucher et~al.(2013)Boucher, Randall, Artaxo,
  et~al.]{BoucherRandall2013}
{\sc Boucher, O., Randall, D., Artaxo, P., et~al.} (2013), \enquote{Clouds and
  aerosols,} in \emph{Climate Change 2013: The Physical Science Basis.
  Contribution of Working Group I to the Fifth Assessment Report of the
  Intergovernmental Panel on Climate Change}, eds. T.~Stocker, D.~Qin, G.-K.
  Plattner, et~al., pp. 571--657, Cambridge University Press.

\bibitem[Boyd et~al.(2011)Boyd, Parikh, Chu, Peleato, and
  Eckstein]{boyd_distributed_2011}
{\sc Boyd, S., Parikh, N., Chu, E., Peleato, B., and Eckstein, J.} (2011),
  \enquote{Distributed {Optimization} and {Statistical} {Learning} via the
  {Alternating} {Direction} {Method} of {Multipliers},} \emph{Foundations and
  Trends in Machine Learning}, {\bf 3}(1), 1--122.

\bibitem[Corless et~al.(1996)Corless, Gonnet, Hare, Jeffrey, and
  Knuth]{corless_lambertw_1996}
{\sc Corless, R.~M., Gonnet, G.~H., Hare, D. E.~G., Jeffrey, D.~J., and Knuth,
  D.~E.} (1996), \enquote{On the {LambertW} function,} \emph{Advances in
  Computational Mathematics}, {\bf 5}(1), 329--359.

\bibitem[Engle(2002)Engle]{engle2002dynamic}
{\sc Engle, R.} (2002), \enquote{Dynamic conditional correlation: A simple
  class of multivariate generalized autoregressive conditional
  heteroskedasticity models,} \emph{Journal of Business \& Economic
  Statistics}, {\bf 20}(3), 339--350.

\bibitem[Fischer et~al.(2013)Fischer, Beyerle, and Knutti]{FischerBeyerle2013}
{\sc Fischer, E.~M., Beyerle, U., and Knutti, R.} (2013), \enquote{Robust
  spatially aggregated projections of climate extremes,} \emph{Nature Climate
  Change}, {\bf 3}, 1033---1038.

\bibitem[Gibberd and Nelson(2017)Gibberd and Nelson]{GibberdNelson2017}
{\sc Gibberd, A.~J., and Nelson, J.~D.} (2017), \enquote{Regularized estimation
  of piecewise constant gaussian graphical models: The group-fused graphical
  lasso,} \emph{Journal of Computational and Graphical Statistics}, {\bf
  26}(3), 623--634.

\bibitem[Grise et~al.(2013)Grise, Polvani, Tselioudis, Wu, and
  Zelinka]{GrisePolvani2013}
{\sc Grise, K.~M., Polvani, L.~M., Tselioudis, G., Wu, Y., and Zelinka, M.~D.}
  (2013), \enquote{The ozone hole indirect effect: Cloud-radiative anomalies
  accompanying the poleward shift of the eddy-driven jet in the southern
  hemisphere,} \emph{Geophysical Research Letters}, {\bf 40}(14), 3688--3692.

\bibitem[Hallac et~al.(2017)Hallac, Park, Boyd, and Leskovec]{HallacPark2017}
{\sc Hallac, D., Park, Y., Boyd, S., and Leskovec, J.} (2017), \enquote{Network
  inference via the time-varying graphical lasso,} in \emph{Proceedings of the
  23rd ACM SIGKDD International Conference on Knowledge Discovery and Data
  Mining}, KDD '17, pp. 205--213, New York, NY, USA, ACM.

\bibitem[Hansen et~al.(2012)Hansen, Sato, and Ruedy]{hansen_perception_2012}
{\sc Hansen, J., Sato, M., and Ruedy, R.} (2012), \enquote{Perception of
  climate change,} \emph{Proceedings of the National Academy of Sciences}, {\bf
  109}(37).

\bibitem[Harvey et~al.(1994)Harvey, Ruiz, and Shephard]{HarveyRuiz1994}
{\sc Harvey, A., Ruiz, E., and Shephard, N.} (1994), \enquote{Multivariate
  stochastic variance models,} \emph{The Review of Economic Studies}, {\bf
  61}(2), 247--264.

\bibitem[Hu et~al.(2015)Hu, Zeng, and Lin]{hu_dual_2015}
{\sc Hu, Q., Zeng, P., and Lin, L.} (2015), \enquote{The dual and degrees of
  freedom of linearly constrained generalized lasso,} \emph{Computational
  Statistics \& Data Analysis}, {\bf 86}, 13--26.

\bibitem[Huntingford et~al.(2013)Huntingford, Jones, Livina, Lenton, and
  Cox]{huntingford_no_2013}
{\sc Huntingford, C., Jones, P.~D., Livina, V.~N., Lenton, T.~M., and Cox,
  P.~M.} (2013), \enquote{No increase in global temperature variability despite
  changing regional patterns,} \emph{Nature}, {\bf 500}(7462), 327--330.

\bibitem[Kahn et~al.(2007)Kahn, Fishbein, Nasiri, Eldering, Fetzer, Garay, and
  Lee]{KahnFishbein2007}
{\sc Kahn, B.~H., Fishbein, E., Nasiri, S.~L., Eldering, A., Fetzer, E.~J.,
  Garay, M.~J., and Lee, S.-Y.} (2007), \enquote{The radiative consistency of
  atmospheric infrared sounder and moderate resolution imaging
  spectroradiometer cloud retrievals,} \emph{Journal of Geophysical Research:
  Atmospheres}, {\bf 112}(D9).

\bibitem[Kim et~al.(2009)Kim, Koh, Boyd, and Gorinevsky]{KimKoh2009}
{\sc Kim, S.-J., Koh, K., Boyd, S., and Gorinevsky, D.} (2009),
  \enquote{$\ell_1$ trend filtering,} \emph{SIAM Review}, {\bf 51}(2),
  339--360.

\bibitem[Monti et~al.(2014)Monti, Hellyer, Sharp, Leech, Anagnostopoulos, and
  Montana]{MontiHellyer2014}
{\sc Monti, R.~P., Hellyer, P., Sharp, D., Leech, R., Anagnostopoulos, C., and
  Montana, G.} (2014), \enquote{Estimating time-varying brain connectivity
  networks from functional {MRI} time series,} \emph{NeuroImage}, {\bf 103},
  427--443.

\bibitem[Myers et~al.(2018)Myers, Mechoso, and DeFlorio]{MyersMechoso2018}
{\sc Myers, T.~A., Mechoso, C.~R., and DeFlorio, M.~J.} (2018),
  \enquote{Importance of positive cloud feedback for tropical atlantic
  interhemispheric climate variability,} \emph{Climate Dynamics}, {\bf
  51}(5-6), 1707--1717.

\bibitem[Nishihara et~al.(2015)Nishihara, Lessard, Recht, Packard, and
  Jordan]{NishiharaLessard2015}
{\sc Nishihara, R., Lessard, L., Recht, B., Packard, A., and Jordan, M.}
  (2015), \enquote{A general analysis of the convergence of admm,} in
  \emph{Proceedings of the 32nd International Conference on Machine Learning},
  eds. F.~Bach and D.~Blei, vol.~37, pp. 343--352, PMLR.

\bibitem[Parikh and Boyd(2014)Parikh and Boyd]{parikh_proximal_2014}
{\sc Parikh, N., and Boyd, S.} (2014), \enquote{Proximal {Algorithms},}
  \emph{Foundations and Trends{\textregistered} in Optimization}, {\bf 1}(3),
  127--239.

\bibitem[Rhines and Huybers(2013)Rhines and Huybers]{rhines_frequent_2013}
{\sc Rhines, A., and Huybers, P.} (2013), \enquote{Frequent summer temperature
  extremes reflect changes in the mean, not the variance,} \emph{Proceedings of
  the National Academy of Sciences}, {\bf 110}(7), E546--E546.

\bibitem[Schreier et~al.(2010)Schreier, Kahn, Eldering, Elliott, Fishbein,
  Irion, and Pagano]{SchreierKahn2010}
{\sc Schreier, M., Kahn, B., Eldering, A., Elliott, D., Fishbein, E., Irion,
  F., and Pagano, T.} (2010), \enquote{Radiance comparisons of modis and airs
  using spatial response information,} \emph{Journal of Atmospheric and Oceanic
  Technology}, {\bf 27}(8), 1331--1342.

\bibitem[Screen(2014)Screen]{Screen2014}
{\sc Screen, J.~A.} (2014), \enquote{Arctic amplification decreases temperature
  variance in northern mid- to high-latitudes,} \emph{Nature Climate Change},
  {\bf 4}, 577---582.

\bibitem[Staten et~al.(2016)Staten, Kahn, Schreier, and
  Heidinger]{StatenKahn2016}
{\sc Staten, P.~W., Kahn, B.~H., Schreier, M.~M., and Heidinger, A.~K.} (2016),
  \enquote{Subpixel characterization of {HIRS} spectral radiances using cloud
  properties from {AVHRR},} \emph{Journal of Atmospheric and Oceanic
  Technology}, {\bf 33}(7), 1519--1538.

\bibitem[Tibshirani(2014)Tibshirani]{Tibshirani2014}
{\sc Tibshirani, R.~J.} (2014), \enquote{Adaptive piecewise polynomial
  estimation via trend filtering,} \emph{Annals of Statistics}, {\bf 42},
  285--323.

\bibitem[Tibshirani and Taylor(2011)Tibshirani and
  Taylor]{TibshiraniTaylor2011}
{\sc Tibshirani, R.~J., and Taylor, J.} (2011), \enquote{The solution path of
  the generalized lasso,} \emph{Annals of Statistics}, {\bf 39}(3), 1335--1371.

\bibitem[Tibshirani and Taylor(2012)Tibshirani and
  Taylor]{tibshirani_degrees_2012}
{\sc Tibshirani, R.~J., and Taylor, J.} (2012), \enquote{Degrees of freedom in
  lasso problems,} \emph{The Annals of Statistics}, {\bf 40}(2), 1198--1232.

\bibitem[Trenberth et~al.(2014)Trenberth, Zhang, Fasullo, and
  Taguchi]{TrenberthZhang2014}
{\sc Trenberth, K.~E., Zhang, Y., Fasullo, J.~T., and Taguchi, S.} (2014),
  \enquote{Climate variability and relationships between top-of-atmosphere
  radiation and temperatures on earth,} \emph{Journal of Geophysical Research:
  Atmospheres}, {\bf 120}(9), 3642--3659.

\bibitem[Vasseur et~al.(2014)Vasseur, DeLong, Gilbert, Greig, Harley, McCann,
  Savage, Tunney, and O{\textquoteright}Connor]{VasseurDeLong2014}
{\sc Vasseur, D.~A., DeLong, J.~P., Gilbert, B., Greig, H.~S., Harley, C.
  D.~G., McCann, K.~S., Savage, V., Tunney, T.~D., and
  O{\textquoteright}Connor, M.~I.} (2014), \enquote{Increased temperature
  variation poses a greater risk to species than climate warming,}
  \emph{Proceedings of the Royal Society of London B: Biological Sciences},
  {\bf 281}(1779).

\bibitem[Wang et~al.(2016)Wang, Sharpnack, Smola, and
  Tibshirani]{WangSharpnack2016}
{\sc Wang, Y.-X., Sharpnack, J., Smola, A.~J., and Tibshirani, R.~J.} (2016),
  \enquote{Trend filtering on graphs,} \emph{Journal of Machine Learning
  Research}, {\bf 17}(105), 1--41.

\bibitem[Wielicki et~al.(2013)Wielicki, Young, Mlynczak, Thome, Leroy, Corliss,
  Anderson, Ao, Bantges, Best, et~al.]{WielickiYoung2013}
{\sc Wielicki, B.~A., Young, D., Mlynczak, M., Thome, K., Leroy, S., Corliss,
  J., Anderson, J., Ao, C., Bantges, R., Best, F., et~al.} (2013),
  \enquote{Achieving climate change absolute accuracy in orbit,} \emph{Bulletin
  of the American Meteorological Society}, {\bf 94}(10), 1519--1539.

\bibitem[Zeng et~al.(2017)Zeng, Hu, and Li]{zeng_geometry_2017}
{\sc Zeng, P., Hu, Q., and Li, X.} (2017), \enquote{Geometry and {Degrees} of
  {Freedom} of {Linearly} {Constrained} {Generalized} {Lasso},}
  \emph{Scandinavian Journal of Statistics}, {\bf 44}(4), 989--1008.

\end{thebibliography}

\appendix

\section{PDIP for $\ell_1$ Trend Filtering of variance}
\label{sec:app_l1tf_var}

In this appendix we provide more details on how to solve the optimization problem with the objective specified in \autoref{eq:l1tf_var} using PDIP. The objective function is convex but not differentiable. Therefore, to be able to use PDIP we first need to derive the dual of this problem. We note that this is a generalized LASSO problem \citep{TibshiraniTaylor2011}. The dual of a generalized LASSO with the objective $f(x)+\lambda \norm{ Dx }_1$ is:  
\begin{align}
\min_\nu&\quad f^*(-D^\top\nu) & \mbox{s.t.}&\quad \norm{ \nu }_\infty \le \lambda
\end{align}
where $f^*(\cdot)$ is the Fenchel conjugate of $f$: $f^*(u)=\max_x u^\top x-f(x)$. It is simple to show that for the objective function of Equation \ref{eq:l1tf_var}
\begin{equation}
f^*(u)=\sum_t (u_t-1)\log\frac{y_t^2}{1-u_t} + u_t-1.
\label{eq:conj}
\end{equation}

Each iteration of PDIP involves computing a search direction by taking a Newton step for the system of nonlinear equations $r_w(v,\mu_1,\mu_2)=0$, where $w>0$ is a parameter and
\begin{equation}
\begin{aligned}
  &r_w(v,\mu_1,\mu_2)
  :=
	\begin{bmatrix}
	r_{dual}\\
	r_{cent}	
	\end{bmatrix}=
  \begin{bmatrix}
    \nabla f^*(-D^\top v) + \mu_1 - \mu_2\\
    -\mu_1(v-\lambda\one)+\mu_2(v + \lambda\one) -w^{-1}\one
  \end{bmatrix}
\end{aligned}
\label{eq:resid}
\end{equation}
where $\mu_1$ and $\mu_2$ are dual variables for the $\ell_\infty$ constraint. Let $A=[\nabla r_{dual}^\top , \nabla r_{cent}^\top]^\top$. The newton step takes the following form
\begin{equation}
r_w(v,\mu_1,\mu_2)+A
\begin{bmatrix}
	\nabla v\\
	\nabla \mu_1\\
	\nabla \mu_2	
	\end{bmatrix}= 0
\label{eq:newton_step}
\end{equation}
with
\begin{equation}
A=
\begin{bmatrix}
	\nabla^2 f^*(-D^\top v) & I & -I\\
	-\mathbf{diag(\mu_1)}\one & -v+\lambda\one & \mathbf{0}\\
	\mathbf{diag(\mu_2)}\one & v+\lambda\one & \mathbf{0}
	\end{bmatrix}.
\label{eq:delta_r}
\end{equation}
Therefore, to perform the Newton step we need to compute $\nabla
f^*(-D^\top v)$ and $\nabla^2 f^*(-D^\top v)$. 
It is straightforward to show that
\begin{align}
\nabla f^*(-D^\top v) &=  -\nabla_u f^*(u) D^\top ,\\
u&=-D^\top v, \\
 (\nabla_u f^*(u))_j&=\log\bigg(\frac{y_j^2}{1-u_j}\bigg), \\
\nabla^2 f^*(-D^\top v)&=D\nabla_u^2 f^*(u)D^\top,\\  
(\nabla_u^2 f^*(u))_j&=\mathbf{diag}\bigg(\frac{1}{1-u_j}\bigg).
\end{align}

Having computed the conjugate function and its gradient and Jacobian,
now we can use a number of convex optimization software packages which
have an implementation of PDIP to solve the optimization problem with
the objective function \autoref{eq:l1tf_var}. We chose the python API of
the \texttt{cvxopt} package~\citep{andersen_cvxopt:_2013}.

\section{PDIP Update in~\autoref{alg:conADMM}}
\label{sec:app_consADMM}

In this section we give more details on performing the $x$-update step in Algorithm 1. We need to solve the following optimization problem:
\begin{align}
\hat{x} &:=\argmin_{x} \bigg( \sum_{j=1}^{n_b} (x_j +
                y_j^2e^{-x_j})
  + (\rho/2) \lVert x-\tilde{z} + u \lVert_2^2 + \Lambda^\top |D x| \bigg)
\label{eq:x_update_opt}
\end{align}
where $n_b$ is the number of local variables in each sub-cube in \autoref{fig:data_cube}, and for ease of notation we have dropped the subscript $i$ and superscript $m$. 

The matrix $D$ has the following form:  $D=[D_{temp}|D_{spat}]$. The matrix $D_{temp}$ is the following block-diagonal matrix and corresponds to the temporal penalty: 
\begin{equation}
D_{temp}=\begin{bmatrix}
D_t &  & \\ 
& \ddots & \\
&  & D_t
\end{bmatrix},
\label{eq:d_t_matrix}
\end{equation}
where $D_t$ was first introduced in Section 2 of the main text and has the following form:
\begin{equation}
 D_t=\begin{bmatrix}
 1 & -2 & 1 &  &  &  &\\ 
 & 1 & -2 & 1 &  &0  &\\ 
 &  &  & \ddots &  &  &\\ 
 & 0 & & 1 & -2 & 1 &  \\ 
 &  &  &   & 1 & -2 & 1 
 \end{bmatrix}.
\label{eq:d_matrix}
\end{equation}

The number of the diagonal blocks in $D_{temp}$ is equal to the grid
size $n_r \times n_c$. Each row of the matrix $D_{spat}$ corresponds
to one spatial constraint in Equation \eqref{eq:l1tf_var_st} in the
text. For example, the first $T$ rows correspond to
$|h_{11t}-h_{21t}|$ for $t=1,...,T$, the next $T$ rows correspond to
$|h_{11t}-h_{12t}|$, and so on.

This optimization problem, is again a generalized LASSO problem with
$f(x)=\sum_{j=1}^{n_b} (x_j + y_j^2e^{-x_j}) + (\rho/2) \lVert
x-\tilde{z} + u \lVert_2^2$.  

As it was explained in Appendix A, the dual of this optimization
problem is: $\min_\nu f^*(-D^\top\nu)$ with the constraints $|\nu_k|
\le \Lambda_k$. To use PDIP we first need to compute the conjugate
function $f^*(\cdot)$. We have: 
\begin{equation}
\begin{aligned}
f^*(\xi)  = \max_x  \xi^\top x - f(x) =  \max_x \sum_{j=1}^{n_b} (\xi_jx_j - x_j - y_j^2e^{-x_j} -
(\rho/2)(x_j-\tilde{z}_j+u_j)). 
\end{aligned}
\label{eq:conjugate}
\end{equation}
Setting the derivative of the terms inside the summation to 0, we obtain:
\begin{equation}
\xi_j-y_j^2e^{-x_j^*}-\rho x_j^* + \rho (\tilde{z}_j-u_j)=0,
\label{eq:x_start}
\end{equation}
where $x^*$ is the maximizer in \ref{eq:conjugate}. Then, it
can be shown that $x_j^*$ which satisfies \eqref{eq:x_start} can be
obtained as follows: 
\begin{align}
x^*_j & = \mathscr{W}\bigg(\frac{y_j^2}{\rho} e^{\phi_j} \bigg) - \phi_j, \\
\phi_j & =\frac{1-\xi_j-\rho(\tilde{z}_j-u_j)}{\rho}.
\end{align}
In this equation, $\mathscr{W}(\cdot)$ is the Lambert W function \cite{corless_lambertw_1996}. Finally, the conjugate function is: $f^*(\xi) = \sum_{j=1}^{n_b} (\xi_jx^*_j - x^*_j - y_j^2e^{-x^*_j} - (\rho/2)(x^*_j-\tilde{z}_j+u_j))$.

To use PDIP, we also need to evaluate $\nabla f^*$ and $\nabla^2 f^*$. First note that $\frac{\partial \mathscr{W}(q)}{\partial q} = \frac{\mathscr{W}(q)}{q(1+\mathscr{W}(q))}$ and $\frac{\partial^2 \mathscr{W}(q)}{\partial q^2} = - \frac{\mathscr{W}^2(q)(\mathscr{W}(q)+q)}{q^2(1+\mathscr{W}(q))^3}$. Using the chain rule we get:
\begin{equation}
\frac{\partial f^*(\xi)}{\partial \xi_j}  =  x^*_j  + \frac{\partial
  x^*_j}{\partial \xi_j} \left[ \xi_j -1 + y_j^2 e^{-x_j^*} + \rho
  (\tilde{z}_j - u_j - x_j^*) \right], 
\label{eq:d_f*_start}
\end{equation}
where we have
\begin{equation}
\frac{\partial x_j^*}{\partial \xi_j}  = \frac{1}{\rho(1+\mathscr{W}((y_j^2/\rho) e^{-\phi_j} ))}.
\label{eq:d_x*_start}
\end{equation}

By some tedious but straightforward computation we can obtain the second derivatives:
\begin{align}
\frac{\partial^2 f^*(\xi)}{\partial \xi_j^2} 
  & =  \frac{\partial
    x_j^*}{\partial \xi_j}
    - \rho \frac{\partial^2
    x_j^*}{\partial
    \xi_j^2} \bigg[ \phi_j
    +x_j^* - \tilde{z}_j +
    u_j \bigg]+ \frac{\partial x_j^*}{\partial \xi_j} \bigg[ 1-y_j^2
    \frac{\partial x_j^*}{\partial \xi_j} e^{-x_j^*} -\rho
    \frac{\partial x_j^*}{\partial \xi_j} \bigg], \label{eq:d2_f*_start}\\
  \frac{\partial^2 x_j^*}{\partial \xi_j^2}  &=
                                               \frac{\mathscr{W}((y_j^2/\rho)
                                               e^{-\phi_j}
                                               )}{\rho^2(1+\mathscr{W}((y_j^2/\rho)
                                               e^{-\phi_j} ))^3} .
\label{eq:d2_x*_start}
\end{align}

\section{Proof of Lemma 1}
\label{sec:proof-lemma-1}

\begin{proof}
  If $f(x)=\sum_k f_k(x_k)$ then $[\prox_{\mu f}(x)]_k =
  \prox_{\mu f_k}(u_k)$. So 
  $[\prox_{\mu f}(u)]_k=\min_{x_k} \,\,
  \mu\big(x_k+y_{k}^2e^{-x_{k}}\big)+\frac{1}{2}  (x_k-u_k)^2.$
  Setting the derivative to 0 and solving for $u_k$ gives the
  result. Similarly, $[\prox_{\rho g}(u)]_\ell=\rho
  \lambda_\ell |z_\ell|+1/2(z_\ell-u_\ell)^2$. This is not differentiable,
  but the solution must satisfy $\rho \cdot \lambda_\ell \cdot \partial
  \big(|z_\ell| \big)=u_\ell-z_\ell$ where $\partial \big(|z_\ell| \big)$ is the
  sub-differential of $|z_\ell|$. The solution is the soft-thresholding
  operator $S_{\rho\lambda_\ell}(u_\ell)$.
\end{proof}

\end{document}